\PassOptionsToPackage{monochrome}{xcolor}
\documentclass[10pt,twocolumn,letterpaper]{article}

\usepackage[datasets]{wacv}

\usepackage{multirow}
\usepackage{amsmath}
\usepackage{amssymb}
\usepackage{graphicx}
\usepackage{xcolor}
\usepackage{booktabs,tabularx,makecell,pifont}
\graphicspath{{./}{./Figures/}}

\newcommand{\cmark}{\ding{51}}
\newcommand{\xmark}{\ding{55}}

\usepackage[pagebackref,breaklinks,hidelinks]{hyperref}

\def\wacvPaperID{2028}
\def\confName{WACV}
\def\confYear{2027}

\title{WZPlanner: Safe End-to-End Path Planning for Autonomous Driving in Work Zones}

\author{
Nishad Sahu \quad Changzhong Qian \quad Guangzhou Cai\\
Shounak Sural \quad Ragunathan (Raj) Rajkumar\\[1mm]
Department of Electrical and Computer Engineering\\
Carnegie Mellon University, Pittsburgh, PA, USA
}

\begin{document}
\maketitle

\begin{abstract}

Work zones are common safety-critical driving environments where lane geometry is frequently altered by temporary traffic control artifacts. Partial or complete lane or road closures in work zones may not be reflected in onboard maps creating both perception and planning challenges for autonomous vehicles (AVs). Existing perception systems and modular AV stacks often fail to generalize well in such settings, due to structural complexities of possible road configurations and in part due to the limited availability of publicly released data with structured geometric supervision. This paper presents the \textbf{\textit{WorkZonePlan}} dataset, a work zone dataset comprising \textbf{149K+ synthetic} and \textbf{5K+ real-world} multimodal samples with \textbf{3D annotations} for lane boundaries, work zone boundaries, and driving trajectory planning options. The dataset also provides 76 closed-loop work-zone scenarios in CARLA, each replayed under three weather conditions to form 228 Bench2Drive-format evaluation routes. We also introduce \textbf{\textit{WAVE}} (\textbf{W}ork-zone-focused \textbf{A}V data generation in \textbf{V}irtual and r\textbf{E}al Environments), a semi-automated pipeline used to create the \textbf{\textit{WorkZonePlan}} dataset. Finally, we propose \textbf{\textbf{\textit{BoundaryFormer}} (BF)}, a transformer-based architecture that jointly predicts lane and work zone boundary polynomials and driving trajectories.

\end{abstract}
\vspace{-12pt}

\begin{figure}[!t]
    \centering
    \includegraphics[width=0.94\linewidth]{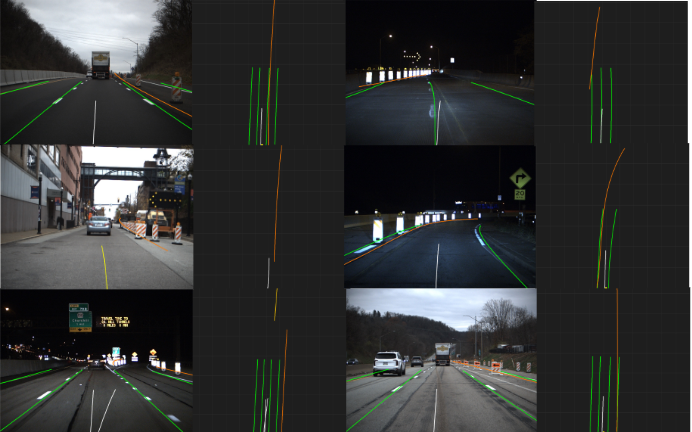}
    \vspace{-2mm}
    \caption{Visualization of camera annotation on some real-world data-points in \textbf{\textit{WorkZonePlan}} dataset along with the bird's-eye view of the annotations on the right. (Grey = Driving trajectory, Green = Lane boundaries, Red/Orange = Work zone boundaries)}
    \label{fig:real_cam}
  \vspace{-6mm}
\end{figure}

{\itshape\noindent
\textbf{\textit{BF}} uses slot attention for boundary prediction. Ablation studies show that a distinct trajectory decoder leveraging boundary slot features improves trajectory prediction substantially compared to predicting trajectories purely within the slot-attention framework. Building on this finding, we introduce \textbf{\textit{BF++}}, a more robust successor with Camera and \textbf{\textit{Camera+LiDAR}} variants. BF++ adds work-zone-specific geometric structure through metric ground-plane encoding, typed boundary/trajectory queries, long-range point anchors, image-space curve refinement, and conservative gated LiDAR fusion. Over the 211 routes common to all four compared models at the evaluation freeze, BF++-Camera and BF++-Camera+LiDAR obtain Driving Scores of 63.0 and 64.4, respectively, compared with 59.3 for SimLingo and 26.1 for TransFuser++ (TF++). BF++ is 40$\times$ smaller than SimLingo and more than 10$\times$ smaller than TF++, while achieving higher Driving Scores. Our results show that a unified end-to-end model prediction of lane, work zone boundaries, and planning trajectories is a promising direction toward safer AV operation in work zones.
Code and dataset are available at \url{https://github.com/Nishad-Sahu/WZPlanner}.
\par}
\vspace{12pt}

\vspace{-6mm}
\section{INTRODUCTION}
\label{sec:introduction}
\vspace{-2mm}

\begin{figure*}
    \centering
    \includegraphics[width=0.74\linewidth]{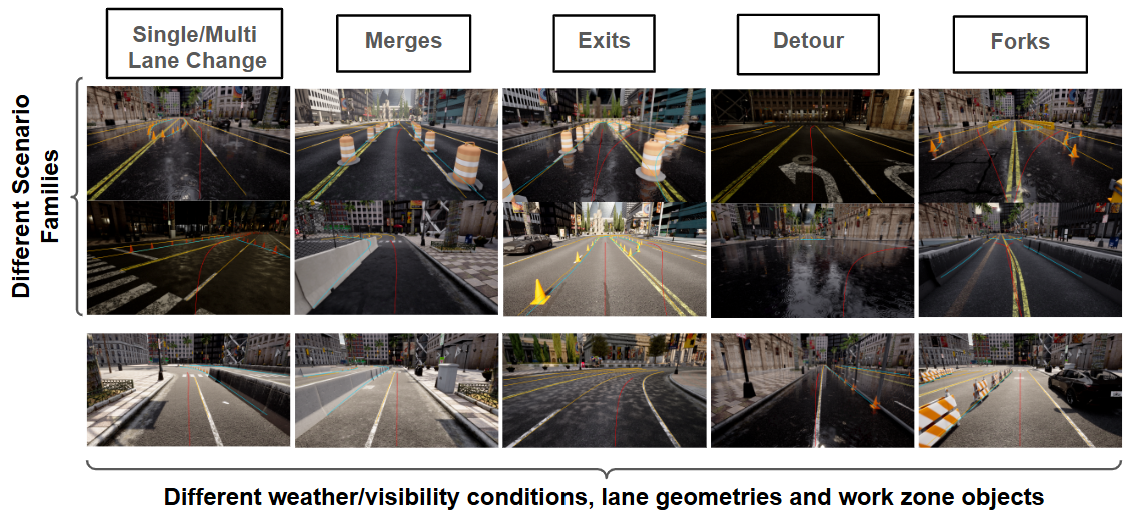}
    \vspace{-3mm}
    \caption{Some synthetic data-points with 3D polynomial annotations in \textbf{\textit{WorkZonePlan}} dataset across different work zone scenarios with different work zone objects in various visibility and weather conditions. (Driving trajectory - Red, work zone boundary - blue, lane boundaries - yellow)}
    \label{fig:syn_master}
  \vspace{-5mm}
\end{figure*}


Autonomous driving technology has made significant strides on structured highways and urban roads, where lane geometry, traffic control and maps are relatively stable. Work zones, however, violate many of these assumptions simultaneously: lanes may narrow or shift, shoulders may temporarily become travel lanes, and detours that are not reflected on maps may appear. Entry and exit ramps may be shifted, and/or intersect a long work zone. In addition, workers and construction vehicles often occupy atypical positions within or near the travel way. These challenges are frequently compounded by adverse weather and reduced visibility. As a result, safe navigation in dynamic and unpredictable environments such as work zones remains an open problem for autonomous vehicles (AVs).

According to the Federal Highway Administration (FHWA), work zones contribute substantially to traffic congestion and safety risk, motivating specialized solutions to ensure reliable AV operation \cite{fhwa_workzone}. FHWA statistics indicate that, in the United States, work zones are associated with approximately one fatality per 4 billion vehicle-miles traveled and per \$112 million in roadway construction expenditures. In 2022, there were 891 work zone-related fatalities, including 742 drivers and passengers and 145 bicyclists and pedestrians \cite{fhwa_workzone}. Therefore, reliable and validated techniques are needed  to detect and navigate a wide range of complex work zone scenarios. Meeting this challenge also needs an exhaustive suite of scenarios for evaluating AV behavior both in simulation, closed test-tracks and in real-world deployments.

\subsection{From Rule-Based Decisions to End-to-End Prediction}

Most prior studies on safe AV navigation in work zones adopt a modular architecture \cite{saferoute,sahu2025towards,shi2021work,cosim}, with distinct modules for work zone perception (e.g., object and boundary detection), lane detection, behavior planning, path planning, and other downstream decision-making tasks. For example, \cite{sahu2025towards} studies a broad range of work zone scenarios defined by a state transportation authority and summarizes behavioral, perception, path-planning, and route-planning requirements in a consolidated, rule-based framework. Their system explicitly switches among \emph{map following}, \emph{lane following}, and \emph{work zone boundary following} depending on the observed context. While such rules-driven designs offer transparency, explainability, explicit controllability and align with operational safety requirements, they do have a key limitation: robust performance depends on correctly identifying the current mode and on engineered switching logic, which may not generalize to the long tail of work zone configurations. Moreover, mathematical or heuristic boundary construction methods such as convex-hull grouping in \cite{shi2021work} can incorrectly merge work zone objects on opposite sides of the roadway into a single boundary, yielding false positives for road blockage even though these work zone objects belong to two distinct boundaries and represent a drivable corridor.

Destination-level navigation and local trajectory prediction are complementary. Airinei et al. use graph-based path generation, explainable augmentation, and curriculum learning for target-conditioned visual navigation \cite{Airinei_2025_ICCV}, while Leordeanu and Paraicu combine visual localization with destination-conditioned trajectory prediction \cite{leordeanu2021driven}. \textbf{\textit{BoundaryFormer}} instead predicts local work-zone boundaries and trajectory options which can be later combined with a higher planning block for navigation to destination. In our CARLA closed-loop evaluation, every baseline model and \textbf{\textit{BoundaryFormer}} receives the same L/R/straight navigation command to select a local trajectory planning options from multiple options based on the goal location.

\begin{table*}[t]
\centering
\caption{Comparison of work-zone datasets and benchmarks. \textit{WorkZonePlan} has 76 closed-loop CARLA scenarios, replayed in three weather conditions to form 228 evaluation routes, together with calibrated 3D lane/work-zone boundaries and trajectory options across adverse conditions in synthetic and real-world environments.}
\label{tab:wz_dataset_comparison}
\renewcommand{\arraystretch}{0.95}
\setlength{\tabcolsep}{0.8pt}
\fontsize{6.5}{7.0}\selectfont

\begin{tabularx}{\textwidth}{@{}l l c c c c c c c c c c c X@{}}
\toprule
\textbf{Dataset} & \textbf{Venue (Year)} &
\thead{Cam} &
\thead{LiDAR} &
\thead{Real} &
\thead{Synth} &
\thead{Calib.\\3D GT} &
\thead{Lane\\Geom.\\GT} &
\thead{WZ\\Contours\\GT} &
\thead{Traj./\\Path GT} &
\thead{Adverse\\Weather/\\Visibility} &
\thead{Closed\\Loop\\WZ Scen.} &
\thead{\#WZ\\Obj.} &
\thead{Dataset Size and description}\\
\midrule

nuScenes~\cite{caesar2020nuscenes} & CVPR '20 &
\cmark & \cmark & \cmark & \xmark & \cmark &
\cmark & \xmark & \xmark & \xmark &
\xmark &
3 &
40k samples; work-zone object classes are sparse \\

BDD100K~\cite{yu2020bdd100k} & CVPR '20 &
\cmark & \xmark & \cmark & \xmark & \xmark &
\xmark & \xmark & \xmark & \cmark &
\xmark &
0 &
100k samples; no work-zone object classes \\

Shi~et~al.~\cite{shi2021work} & ITSC '21 &
\cmark & \cmark & \cmark & \xmark & Partial &
\xmark & \xmark & \xmark & \xmark &
\xmark &
4 &
757 nuScenes samples with WZ labels \\

RZDG~\cite{yan2025framework} & IV '25 &
\cmark & \cmark & \cmark & \cmark & \cmark &
\xmark & \xmark & \cmark & \xmark &
\xmark &
2 &
8,591 synthetic + 1,357 real frames \\

ROADWork~\cite{ghosh2025roadwork} & ICCV '25 &
\cmark & \xmark & \cmark & \xmark & \makecell{Partial\\(recon.)} &
\xmark & \xmark & \makecell{Partial\\(recon.)} & \cmark &
\xmark &
15 &
4,375 videos; 9,650 keyframes; 129k path images; 5,430 with reconstructed 3D trajectories \\

WorkZone3D~\cite{Sural_2026_WACV} & WACV '26 &
\cmark & \cmark & \cmark & \xmark & \cmark &
\xmark & \xmark & \xmark & \cmark &
\xmark &
3 &
21,650 real samples \\

\midrule
\textbf{\textit{WorkZonePlan} (Ours)} &  &
\cmark & \cmark & \cmark & \cmark & \cmark &
\cmark & \cmark & \cmark & \cmark &
\cmark &
\textbf{9} &
\textbf{149,478} synthetic + \textbf{5,178} WorkZone3D-derived real samples (\textbf{154,656} total) \\
\bottomrule
\end{tabularx}

\vspace{0.5mm}
\parbox{\textwidth}{\tiny \textbf{Note:} ``Designed WZ Scen.'' counts purpose-built layouts, not frames, routes, or observed sites (e.g., ROADWork's $>$5K work zones); N/R = no comparable count reported. ``Partial (recon.)'' = reconstructed rather than direct calibrated 3D labels. WZ = work zone; GT = ground truth; Calib. = calibrated; Geom. = geometry; Traj. = trajectory; Obj. = objects.}
\vspace{-4mm}
\end{table*}

The relative strengths and weaknesses of a layered architecture \textit{vs} a consolidated approach motivate a complementary question explored in this paper: \emph{Can a single neural network directly predict feasible driving trajectory options from perception inputs in work zones?} If so, a conventional route planner (which we treat as standard) can select among the network's predicted trajectory candidates based on route-level objectives, while the perception-to-trajectory network handles local geometry and safety constraints. This paper makes the following contributions:

\begin{itemize}
    \item \textbf{\textit{WAVE}} (\textbf{W}ork-zone-focused \textbf{A}V data generation in \textbf{V}irtual and r\textbf{E}al Environments): a scalable, semi-automated pipeline that generates work-zone scenarios and produces \textbf{3D annotations} for lane boundaries, work-zone boundaries, and driving trajectories in simulation and on real data.
    \item \textbf{\textit{WorkZonePlan} dataset:} We introduce \textbf{149,478 synthetic} and \textbf{5,178 real} samples, for \textbf{154,656 samples} in total (Figs. \ref{fig:syn_master} and \ref{fig:real_cam}); the latter are an exact open-source WorkZone3D subset \cite{Sural_2026_WACV} with added boundary and trajectory labels. The closed-loop benchmark contains 76 work-zone scenarios distributed across CARLA Towns01--05 and Town10 and replayed under three weather conditions, yielding 228 evaluation routes. The dataset covers work zone scenarios designed manually using transportation-authority-inspired layouts spanning lane changes, merges, exits, detours, and forks, nine work-zone object types, and nine weather/visibility combinations (Day/Dusk/Night $\times$ No/Medium/Heavy rain).
    \item \textbf{BoundaryFormer (BF):} We propose a transformer-based architecture that jointly predicts lane boundaries, work-zone boundaries, and driving trajectory options from perception inputs, using slot attention for structured scene parsing and a dedicated trajectory decoder for global, long-horizon reasoning. This can be used as a baseline model for the \textbf{\textit{WorkZonePlan}} dataset by researchers.
    \item \textbf{BoundaryFormer++ (BF++):} Building on the superior \textbf{\textit{BF-DistinctPlanner}} result, we develop camera-only and camera+LiDAR model with metric, type-aware, and topology-oriented inductive biases for work-zones.
    \item \textbf{Experimental evaluation and ablation study:} We evaluate BF++ in open loop and on a closed-loop benchmark of 228 routes spanning six towns and three weather conditions, with external driving baselines and separate camera versus camera+LiDAR results. The four-model comparison uses the 211 routes common to all methods at the evaluation freeze; complete 228-route results are reported in the supplementary material.
\end{itemize}
\vspace{-2mm}
\subsection{Problem Statement}
\vspace{-1mm}
We consider the task of jointly predicting (i) lane boundaries, (ii) work zone boundaries, and (iii) driving trajectory options through the work zone directly from perception. BF uses a monocular RGB image, whereas BF++ is evaluated in both RGB-only and RGB+LiDAR forms. The outputs are structured geometric entities represented as parametric polynomials (third-order polynomial coefficients with valid spatial ranges).

\section{Related Work}
\label{sec:related_work}

\subsection{Work-zone Datasets and Benchmarks}
Work zones have recently gained attention as a common yet under-explored operational domain in autonomous driving. Common AV datasets such as nuScenes \cite{caesar2020nuscenes} and BDD100K \cite{yu2020bdd100k} include work-zone-related object classes (e.g., cones and barriers), but these scenarios are relatively sparse and typically lack structured supervision for work-zone geometry \cite{shi2021work}. The ROADWork dataset \cite{ghosh2025roadwork} provides a relatively large-scale real-world benchmark for recognizing and analyzing work zones, including annotated images and videos from thousands of work zones across multiple cities, and tasks such as object/sign detection and pathway estimation. ROADWork notes that foundation models and standard perception pipelines may fail in these scenarios, underscoring the need for dedicated datasets and models. However, ROADWork primarily provides 2D annotations for work-zone objects/signs and driving pathways; its 3D pathway labels are derived via 3D reconstruction rather than directly captured using calibrated LiDAR and GPS positioning sensors. The RZDG dataset \cite{yan2025framework} provides both real and simulated work zone data but is only limited to 2 work zone object classes. All these datasets lack varying weather and visibility conditions and are mostly collected under good operating conditions. The WorkZone3D dataset \cite{Sural_2026_WACV} provides camera+Lidar work zone data in bad weather conditions like day, night, rain, snow but lacks trajectory annotations for path planning and has only three work zone object classes. 

\begin{figure*}
    \centering
    \includegraphics[width=0.95\linewidth]{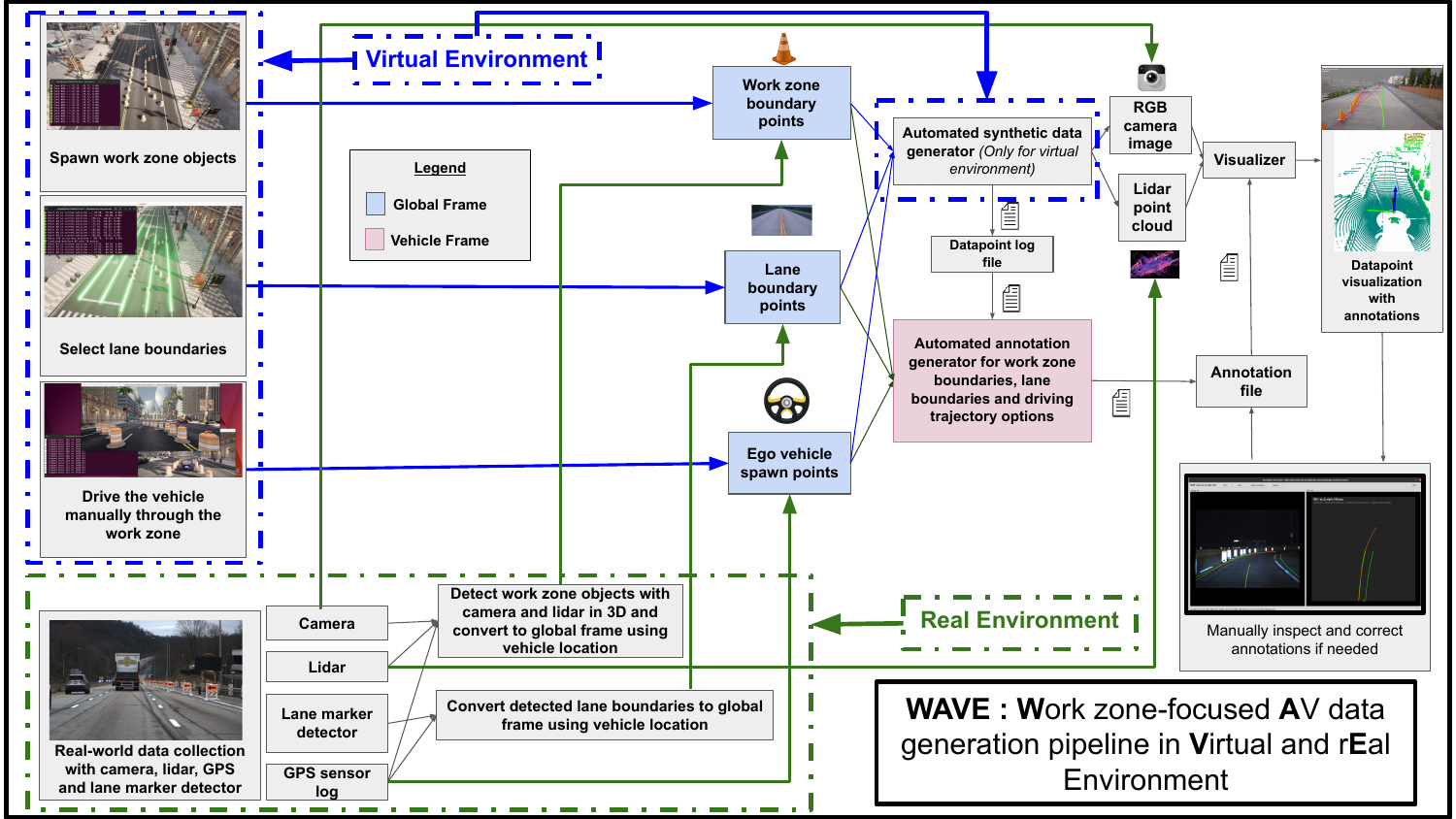}
    \vspace{-2mm}
    \caption{Overview of \textbf{\textit{WAVE}}: \textbf{W}ork zone-focused \textbf{A}V data generation in \textbf{V}irtual and r\textbf{E}al Environments }
    \label{fig:WAVE}
  \vspace{-5mm}
\end{figure*}  


\textbf{\textit{WorkZonePlan}}, the dataset presented in this paper is complementary to the existing datasets. Table \ref{tab:wz_dataset_comparison} presents a comparison of our dataset with other state-of-the-art autonomous driving and work zone datasets. It has a variety of adverse weather and visibility conditions across different work zone scenarios. It also provides annotations for lane geometry and work-zone boundary contours, both which are critical for predicting feasible driving trajectories in reconfigured road layouts. In other words, our focus is on \emph{fused (or one-shot) geometric supervision} for \emph{boundaries and trajectories}. The dataset has \textbf{3D annotations} and calibrated sensor metadata, enabling precise evaluation of boundary localization and trajectory feasibility under controlled scenario attributes. The dataset also combines \textbf{large-scale synthetic} \textbf{data} generation for broad coverage supplemented by \textbf{real-world AV data} to support domain adaptation and transfer experiments. Finally, its 76 Bench2Drive-style~\cite{jia2024bench2drive} work-zone scenarios are based on transportation-authority specifications and are replayed in three weather conditions to form 228 closed-loop routes, capturing complex, operationally realistic work-zone configurations.

The dataset differs from standard lane datasets by explicitly representing \emph{work-zone boundaries} as distinct entities, enabling models to reason about temporary traffic-control geometry that may not align with painted lane markings.

\subsection{Boundary Detection and Trajectory Prediction}
Lane marker detection has evolved from hand-crafted feature pipelines to deep learning approaches that regress polynomials or splines in the image plane and, more recently, to estimating 3D lane geometry \cite{bi2025lane}. Recent state-of-the-art methods increasingly also leverage transformers for 3D lane prediction \cite{luo2023latr,pittner2025sparselanestp}. While several studies detect work-zone objects \cite{cosim,shi2021work,ghosh2025roadwork}, comparatively few address the prediction of \emph{work-zone boundary contours}. For instance, \cite{shi2021work} proposes inferring work-zone contours from detected work-zone objects.

Trajectory prediction is widely studied in autonomous driving \cite{huang2022survey}, where multiple future trajectories are plausible due to route choice, interactions, and topology. However, many trajectory predictors assume stable lane geometry and/or rely on HD maps. Work zones however violate these assumptions by introducing additional constraints and an altered set of feasible paths. The need to detect such changes motivates \emph{joint boundary-and-driving-trajectory prediction}, where a model can infer both the drivable corridor boundaries and candidate trajectories directly from perception. To the best of our knowledge, no prior work jointly addresses lane-boundary detection, work-zone boundary detection, and multi-modal trajectory prediction. We utilize a unified transformer-based architecture with attention to accomplish this objective. 

\section{\textbf{\textit{WAVE}}: Dataset Generation Pipeline }
\label{sec:wave}
\begin{figure}
    \centering
    \includegraphics[width=0.75\linewidth]{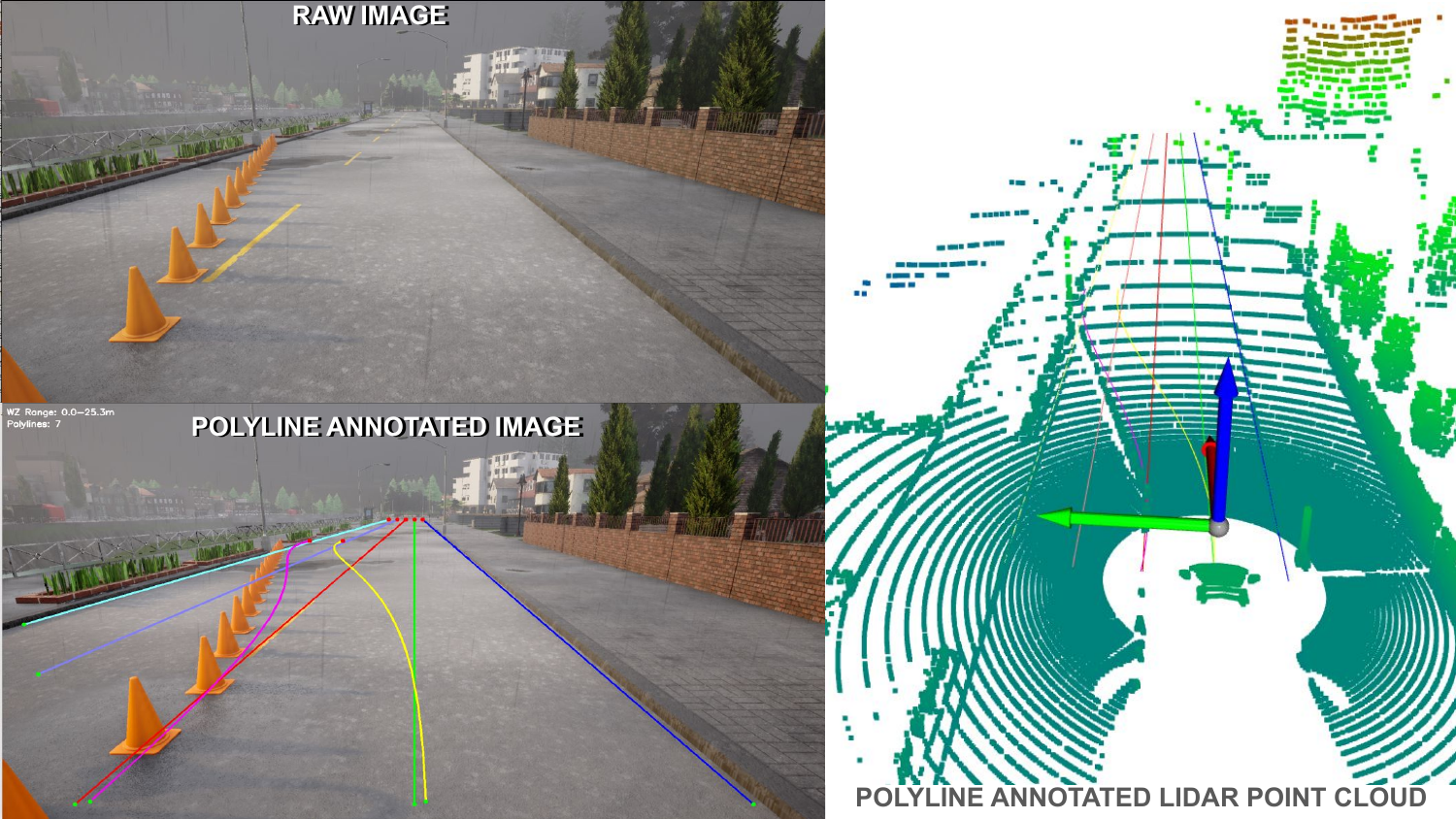}
    \vspace{-2mm}
    \caption{Visualization of Camera and Lidar annotation on a synthetic datapoint in \textbf{\textit{WorkZonePlan}} dataset.}
    \label{fig:syn_cam_lidar}
  \vspace{-8mm}
\end{figure}

We present \textbf{\textit{WAVE}} (\textbf{W}ork-zone-focused \textbf{A}V data generation in \textbf{V}irtual and r\textbf{E}al Environments), a semi-automated pipeline that enables scalable generation of work zone driving data on public roadways as well as in a virtual environment like CARLA. It generates calibration-consistent 3D geometric annotations for work zone objects, work zone boundaries, lane boundaries and possible driving trajectory options (see Figs.~\ref{fig:syn_master},\ref{fig:syn_cam_lidar}). 

The overview of the \textbf{\textit{WAVE}} data generation pipeline is presented in Fig. \ref{fig:WAVE}. \textbf{\textit{WAVE}} combines lightweight human-in-the-loop specification with automated data capture and labeling. In the synthetic world an operator first uses a GUI in the CARLA spectator view to (i) spawn work zone assets, and (ii) select relevant lane boundaries. Then, the operator manually drives the ego vehicle through the work zone to record a representative path. These interactions provide three inputs in the global frame: work zone boundary points, lane boundary points, and ego trajectory waypoints.

Given these inputs, \textbf{\textit{WAVE}} executes an automated synthetic data generator that replays the scene and captures synchronized RGB images and LiDAR point clouds under diverse environment settings: time (day, dusk, night), weather (rain/no rain) and variety of work zone objects (cones, barriers, barrels, fences etc.). All sensor data are logged alongside the required calibration metadata. Next, an automated annotation module transforms global-frame boundary points into an ego-centric coordinate frame and compares the actual 3D depth of the boundary points with the depth of the corresponding pixels in the camera image. If both distances are not within a certain threshold (say 25 cm) then it is highly likely that the annotation point is occluded from the RGB camera perspective and hence is dropped. Then, based on the filtered data, the script generates 3D annotations for lane boundaries, work zone boundaries, and driving trajectory options. Finally, \textbf{\textit{WAVE}} produces a compact annotation file per datapoint and a visualization overlay (as shown in Fig. \ref{fig:syn_cam_lidar}) for rapid quality assurance, enabling efficient iteration over scenario variations and large-scale dataset creation. For real data, \textbf{\textit{WAVE}} uses a 5,178-frame WorkZone3D subset \cite{Sural_2026_WACV}, retaining its synchronized sensors, GPS-RTK/vehicle state, calibration, and 3D boxes. It converts inherited boxes and lane detections to a common global frame to add work-zone/lane boundaries and trajectories, with GUI-based inspection and correction.

\vspace{-2mm}
\section{The \textbf{\textit{WorkZonePlan}} Dataset}
\label{sec:WorkZonePlan}

\textbf{Scenario Families and Transportation Authority-Specified Designs}:
\textbf{\textit{WorkZonePlan}} dataset has manually designed work-zone scenario layouts organized into families based on transportation-authority specifications. These include single/multi-lane changes, merges, exits, detours and forks. The members of each scenario family are grouped together because they induce unique changes in feasible driving corridors and a similar behavior maybe required from the AV.
\\
\textbf{{Work Zone Object Library}}: We use a curated set of \textbf{nine} work zone/boundary objects including cones, barrels, horizontal panels, fences, zipper barriers, vertical flat panels and multiple concrete barrier variants (Supplementary Fig.~S2). These assets are procedurally placed to form work zone boundaries consistent with the intended scenario topology.
\\
\textbf{{Weather and Visibility Conditions}}: To evaluate robustness under adverse perception conditions, \textbf{\textit{WorkZonePlan}} includes \textbf{nine} weather/visibility combinations formed by Day/Dusk/Night lighting crossed with No/Medium/Heavy rain. The real-world data has  day/night and rain/no-rain. 

\textbf{{Sensors and annotations}}: Each frame includes synchronized camera/LiDAR data and \textbf{3D annotations} represented as third-order polynomials with valid ranges and objectness. The real subset retains WorkZone3D's six 16-channel VLP-16 LiDARs and front RGB camera: $1024{\times}768$ ($25.3^\circ{\times}17.3^\circ$). Camera/LiDAR acquisition is 30/20 Hz or 10/10 Hz. We use precise timestamps and 6-DoF UTM state support alignment, with OpenCV, cross-path, and object-guided calibration as discussed in \cite{Sural_2026_WACV}. We add lane and work-zone boundaries along with driving trajectories for training and evaluation.

\textbf{{Dataset composition}}: \textbf{\textit{WorkZonePlan}} has \textbf{137,749-frame} Town01/Town10 synthetic corpus and adds a leakage-resistant \textbf{11,729-frame} held-out set from Town02,03,04,05 (2,797/4,945/1,792/2,195 frames, respectively). The resulting release contains \textbf{149,478 synthetic} CARLA frames and \textbf{5,178 real-world} frames. The original corpus contains 80K+ Town01 and 50K+ Town10 frames and provides the in-distribution data, while Town02--05 are reserved for out-of-distribution evaluation. The closed-loop benchmark uses 12 base scenarios in each of Towns01--05 and 16 additional in-distribution scenarios in Town10. Replaying every scenario under clear, fog, and storm conditions gives 76 base scenarios and 228 evaluation routes. Supplementary Sec.~S6 and Table~S4 provide the town-wise frame, modality, split, and closed-loop distribution; Supplementary Sec.~S5 distinguishes the 211-route common comparison from the complete 228-route records. The real samples are derived from WorkZone3D; all source modalities and 3D boxes are retained, and our structured boundary and driving trajectory labels are added. We use synthetic data for training and ablations and real data for fine-tuning and domain transfer.


\section{The \textbf{\textit{BoundaryFormer}} Architecture}
\label{sec:boundaryformer}
\begin{figure}
    \centering
    \includegraphics[width=0.94\linewidth]{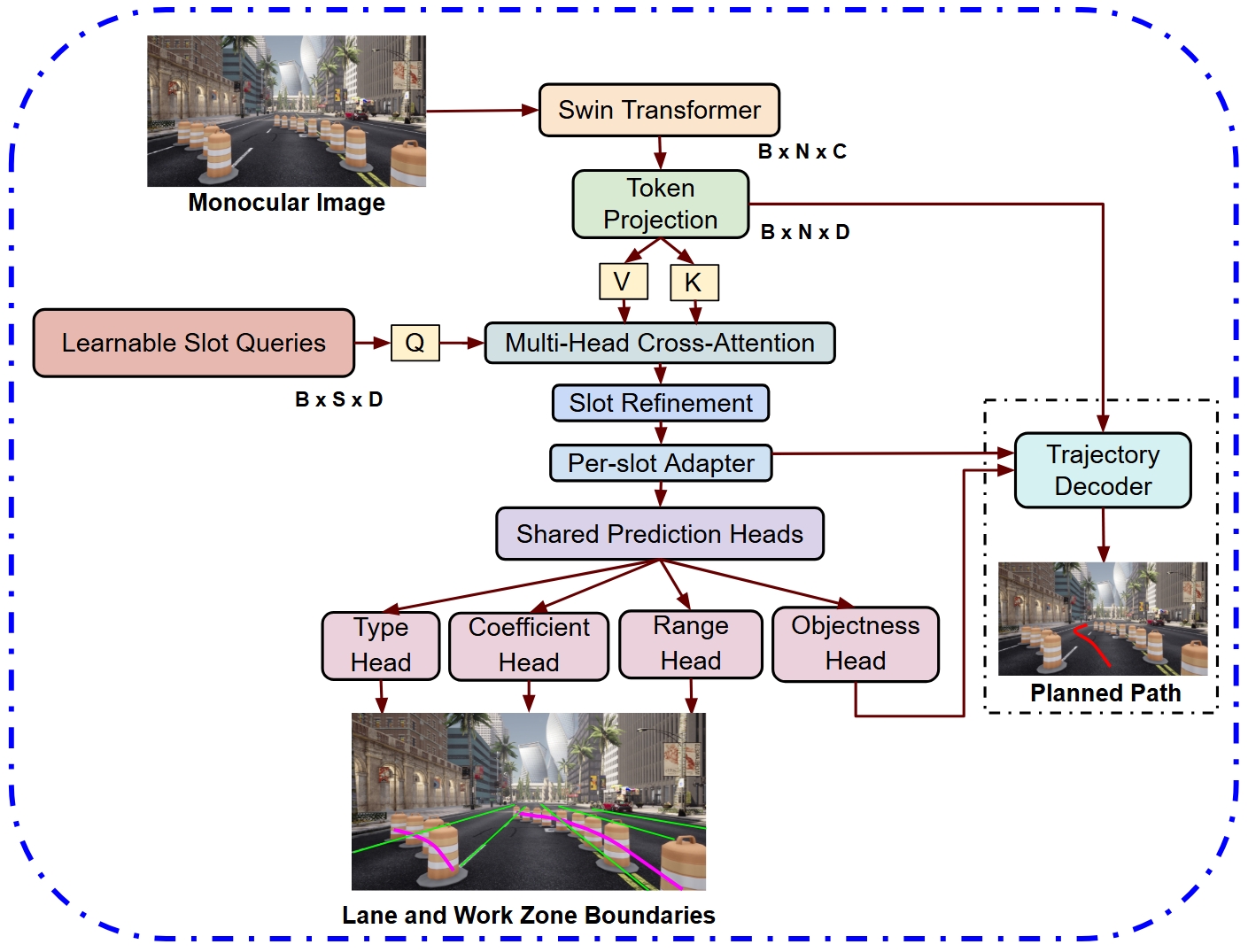}
    \vspace{-2mm}
    \caption{Overview of the \textbf{\textit{\textbf{\textit{BoundaryFormer}}}} architecture. In the \textbf{\textit{BF-IntegratedPlanner}}, the driving trajectory is one of the outputs of the shared prediction heads. In the \textbf{\textit{BF-DistinctPlanner}}, the shared prediction heads predict only the lane and work zone boundaries, with trajectory options predicted by the trajectory decoder. (B = Batch size, N = No. of spatial tokens, C = Backbone channel dimension, D = Hidden dimension of token projection block and learnable slot queries block, S = No. of slots, Q = Queries, K = Keys and V = Values) }
    \label{fig:BF_arch}
  \vspace{-5mm}
\end{figure}

\vspace{-2mm}
We present a new transformer-based architecture called \textbf{\textit{BoundaryFormer}} that predicts lane boundaries, work zone boundaries, and driving trajectories from a single RGB image using slot attention for structured scene parsing. An overview is given in Fig. \ref{fig:BF_arch}. To keep the nomenclature fixed throughout the paper, \textbf{\textit{BF-IntegratedPlanner}} denotes the integrated shared-slot design, in which trajectories are predicted by the shared slot-attention heads and \textbf{\textit{BF-DistinctPlanner}} denotes the design with a distinct dedicated trajectory decoder (slots predicting only boundaries).

\subsection{Backbone and Token Projection}
\vspace{-2mm}
An RGB image is processed by a pre-trained Swin Transformer backbone to extract spatial tokens. A two-layer token encoder projects these tokens into a shared latent space, stabilizing feature variance and improving interactions between visual tokens and learnable slot queries.

\subsection{Slot Attention for Structured Entities}
\vspace{-2mm}
Learnable slot queries access the projected tokens via multi-head cross-attention to produce a set of slot features intended to correspond to lane boundaries, work zone boundaries, and (in \textbf{\textit{BF-IntegratedPlanner}}) driving trajectories. Slot representations are then refined and passed through lightweight per-slot adapters. Shared prediction heads output object type, third-order polynomial coefficients, spatial range, and objectness scores for each predicted entity.

\begin{figure*}[!t]
\centering
\begin{minipage}[t]{0.60\textwidth}
\centering
\includegraphics[width=\linewidth]{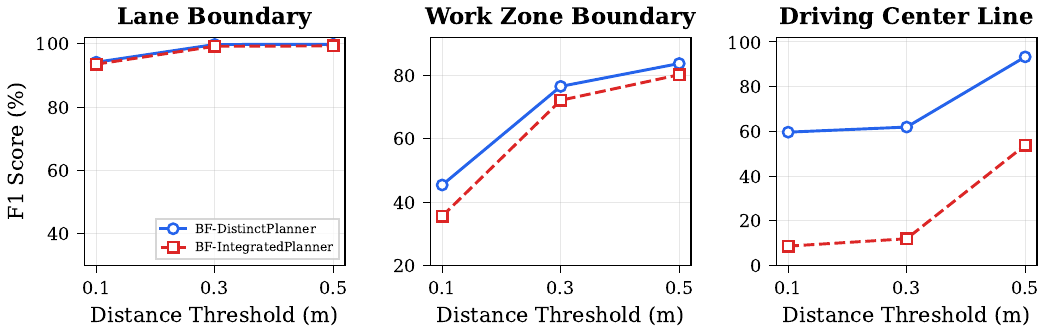}
\vspace{-2mm}
\captionof{figure}{Per-class F1 in Experiment~1 across matching thresholds.
BF-DistinctPlanner is the dedicated-decoder variant (blue circles), whereas
BF-IntegratedPlanner predicts trajectories from shared slots (red squares).}
\label{fig:exp1_f1}
\end{minipage}
\hfill
\begin{minipage}[t]{0.37\textwidth}
\centering
\includegraphics[width=\linewidth]{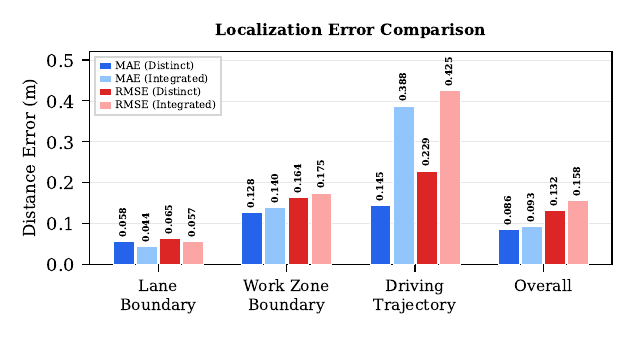}
\vspace{-2mm}
\captionof{figure}{Matched-curve localization errors at the 0.5\,m threshold in
Experiment~1. Distinct denotes the dedicated decoder and Integrated denotes
shared-slot trajectory prediction.}
\label{fig:exp1_localization}
\end{minipage}
\vspace{-3mm}
\end{figure*}

\subsection{Slot-Based Boundary Prediction Heads}

\vspace{-2mm}
Each slot is decoded by four prediction heads. The coefficient head predicts cubic curve parameters for the boundary shape and position. The range head predicts the valid vertical start and end positions of the curve. The type head predicts the boundary class, such as lane boundary or work-zone boundary. The objectness head predicts whether the slot corresponds to a real boundary.

During training, Hungarian matching assigns predicted slots to ground-truth boundaries. Matched pairs supervise the coefficient, range, and type heads, while objectness is trained to distinguish matched boundary slots from unmatched slots.

\subsection{Dedicated Trajectory Decoder in \textbf{\textit{BF-DistinctPlanner}}}
\vspace{-2mm}
Modeling trajectories solely as one of the slots proves to be insufficient for long-horizon reasoning required for driving trajectory because slot attention tends to emphasize local spatial patterns (see Fig.~\ref{fig:exp1_f1}). We therefore introduce a dedicated trajectory decoder operating in parallel with slot-based heads. The decoder aggregates global context via mean pooling of projected tokens and extracts boundary-related cues via objectness-weighted pooling of slot features. These representations are fused using a lightweight MLP to output $K$ trajectory hypotheses (each with polynomial coefficients and spatial range) along with mode probabilities $\pi_k$ for hypotheses $k=1,\ldots,K$; see Supplementary Sec.~S1.2. The $K$ alternatives are generated before route handoff; trajectory-only matching scores coverage, and $\pi_k$ supports selection. This decouples object-level perception from global trajectory reasoning while retaining interpretable slot representations.

\subsection{Training Objective}
\vspace{-2mm}
All \textbf{BF} outputs are jointly optimized using a composite matching-based loss over geometry, range, semantic type, objectness, false positives/negatives, and multimodal trajectory error. Hungarian assignment aligns the variable-size prediction set with ground truth. For details Refer Suppl.Sec.~S1.
\vspace{-2mm}
\section{From BF to BF++: Work-Zone-Specific Robust Prediction}
\label{sec:bfpp}
\vspace{-2mm}
Our experiments in section \ref{sec:experiments_results} shows that BF-DistinctPlanner's dedicated decoder is stronger than treating trajectories as interchangeable shared slots. BF++ builds on this result by assigning boundary and trajectory reasoning explicit metric and semantic roles.

\subsection{Topology-Oriented Design}
\vspace{-2mm}
Temporary devices cause drivable corridors to split, merge, shift, appear, and disappear. BF++ represents this topology using calibrated ground-plane coordinates, non-interchangeable typed queries, 38 longitudinal anchors over 1.5--100\,m with visibility and range, and two-stage curve reprojection into multi-scale image evidence. These components provide task-specific geometric inductive biases.
\vspace{-2.5mm}
\subsection{Camera and Camera+LiDAR Variants}
\vspace{-2mm}
BF++-Camera combines a ConvNeXt-Tiny trunk with a three-layer typed-query decoder (10 lane, 6 work-zone, and 1 trajectory query). The trajectory query emits two independently parameterized, confidence-scored centerlines. BF++-Camera+LiDAR adds only a 1.24M-parameter encoder for projected depth, height, and hit-mask features through a zero-initialized gate. The matched models contain 33.5M and 34.77M parameters. Full architectural, training, and multimodal-scoring details are in Supplementary Secs.~S3.2--S3.4 and S3.6.

\subsection{Closed-Loop Driving and Speed Control}
\label{sec:bfpp_closed_loop}
\vspace{-2mm}
{
During CARLA rollout, the route command selects the highest-confidence compatible BF++ centerline. A shared pure-pursuit controller tracks it with a 6\,m lookahead and 2.85\,m wheelbase. Boundary-midpoint lane snapping is disabled within 25\,m of a detected work zone. A separate 32-channel utility LiDAR, rather than the learned 64-channel LiDAR input, controls speed for both variants. Its fifth-percentile corridor range drives a stopping-distance governor with an 8.9\,m/s nominal speed, 2.0\,m/s$^2$ deceleration, and 3.0\,m buffer. A narrow-corridor range below 2.4\,m commands a full stop. This shared non-PID controller isolates the learned-model comparison. Exact filters, equations, and actuation thresholds are in Supplementary Sec.~S3.5.
}

\section{Experiments and Results}
\label{sec:experiments_results}

\subsection{Methodology and Evaluation Metrics}
\label{sec:eval_metrics}
\vspace{-2mm}
We perform experiments in four stages: (i) the original BF architecture comparison on a CARLA \textit{TOWN 01} 81K synthetic subset, (ii) scaling BF-DistinctPlanner to the original 137K+ Town01/Town10 training subset, (iii) BF synthetic-to-real fine-tuning, and (iv) BF++ camera and camera+LiDAR open-/closed-loop evaluation in CARLA and real-world in-/out-of-distribution settings. The complete 149,478-frame synthetic release adds only the held-out Town02--05 data to the original 137,749-frame Town01/Town10 corpus described in Sec.~\ref{sec:WorkZonePlan}.
Open-loop curve-matching, localization, and trajectory-option
metrics are defined in Supplementary Secs.~S2.1 and S3.6. Closed-loop metric
definitions and aggregation are provided in Supplementary Sec.~S5.1.

\subsection{Experiment 1: \textit{TOWN01} 81K synthetic data architecture comparison}
Following the definitions in Sec. \ref{sec:boundaryformer}, we compare \textbf{\textit{BF-IntegratedPlanner}} (shared-slot trajectory prediction) against \textbf{\textit{BF-DistinctPlanner}} (dedicated trajectory decoder). F1 and localization results are shown in Figs.~\ref{fig:exp1_f1} and \ref{fig:exp1_localization}, respectively. With a random 90/10 training/validation split, both variants nearly saturate lane-boundary localization at moderate thresholds. The driving-centerline results show a much larger gap: \textbf{\textit{BF-DistinctPlanner}} substantially outperforms \textbf{\textit{BF-IntegratedPlanner}} at strict distance thresholds. Work-zone boundary performance also favors \textbf{\textit{BF-DistinctPlanner}}, though by a smaller margin. These results support separating global trajectory reasoning from the local slot outputs and directly motivate the explicit typed trajectory branch in BF++ (Sec.~\ref{sec:bfpp}).

\subsection{Experiment 2: \textbf{\textit{BF-DistinctPlanner}} on Training and Evaluation on 137K+ Synthetic Data}


We scale BF-DistinctPlanner to the original 137K+ synthetic training subset from CARLA \textit{TOWN01} and \textit{TOWN10}, retaining the random 90/10 split. At 0.5\,m, lane boundaries achieve P/R/F1 = 82.2/88.6/85.3 and trajectories 76.7/75.2/75.9; work-zone boundaries remain harder at 58.8/53.1/55.8. See Supplementary Table~S1 (Exp.~2) for threshold and localization results and Fig.~S1 (left) for the 97.38\%-accurate confusion matrix. Its dominant work-zone-to-lane confusion motivates BF++'s non-interchangeable typed queries.

\subsection{Experiment 3: Synthetic-to-Real Fine-Tuning}

We next fine-tune the best Experiment~2 model on all 5,178 real-world samples with the same random 90/10 split. At 0.5\,m, overall P/R/F1 is 60.7/91.9/73.1. Trajectories achieve 62.1/90.5/73.7 with 0.042/0.057\,m MAE/RMSE, lanes 89.8/96.2/92.9, and work-zone boundaries 29.8/81.9/43.7. The easier trajectory geometry relative to the diverse synthetic scenarios reinforces the need for broad synthetic coverage. See Supplementary Table~S1 (Exp.~3) for threshold and localization results and Fig.~S1 (right) for 92.64\% type accuracy, dominant work-zone-to-lane errors, and near-zero trajectory confusion.

\subsection{Experiment 4: BF++ Camera and Camera+LiDAR}
\label{sec:bfpp_results}
\vspace{-2mm}

\subsubsection{In-Distribution Open-Loop Results}
Tables~\ref{tab:bfpp_open_loop_loss_main} and
\ref{tab:bfpp_open_loop_f1_main} report the ID comparison. Similar F1 and route
error indicate saturation on familiar towns (Supplementary Sec.~S3.7).

\begin{table}[!t]
\centering
\caption{BF++ open-loop losses. ID = In Distribution Town01/Town10 validation split and OOD= Out-of-Distribution-Data from Town02,03,04,05.}
\vspace{-2mm}
\label{tab:bfpp_open_loop_loss_main}
\setlength{\tabcolsep}{2.4pt}
\resizebox{\columnwidth}{!}{%
{
\begin{tabular}{llrrrr}
\toprule
\textbf{Split} & \textbf{Model} & \textbf{Total} & \textbf{Obj.} & \textbf{Point} &
\textbf{Route-ADE (m)} \\
\midrule
ID & BF++-Camera & 2.298 & \textbf{0.110} & 2.097 & 0.081 \\
ID & BF++-Camera+LiDAR & \textbf{2.247} & 0.117 & \textbf{2.038} & 0.081 \\
\midrule
OOD & BF++-Camera & 16.443 & 3.125 & \textbf{12.532} & \textbf{0.605} \\
OOD & BF++-Camera+LiDAR & 16.443 & \textbf{3.082} & 12.569 & 0.610 \\
\bottomrule
\end{tabular}
}}
\vspace{-3mm}
\end{table}

\subsubsection{Out-of-Distribution Open-Loop Results}
Tables~\ref{tab:bfpp_open_loop_loss_main} and
\ref{tab:bfpp_open_loop_f1_main} report the OOD comparison. Both variants
degrade similarly, and LiDAR does not improve open-loop generalization in synthetic dataset
(Supplementary Sec.~S3.8).

\begin{table}[!t]
\centering
\caption{BF++ matched-instance F1 at 0.5\,m.}
\vspace{-3mm}
\label{tab:bfpp_open_loop_f1_main}
\setlength{\tabcolsep}{1.8pt}
\resizebox{\columnwidth}{!}{%
{
\begin{tabular}{llrrrr}
\toprule
\textbf{Split} & \textbf{Model} & \textbf{Lane F1} &
\textbf{WZ F1} & \textbf{Traj. F1} & \textbf{Overall F1} \\
\midrule
ID & BF++-Camera & \textbf{0.972} & 0.715 & 0.967 & 0.926 \\
ID & BF++-Camera+LiDAR & 0.971 & \textbf{0.721} & 0.967 & \textbf{0.927} \\
\midrule
OOD & BF++-Camera & \textbf{0.515} & \textbf{0.174} & 0.400$^{\dagger}$ & \textbf{0.447} \\
OOD & BF++-Camera+LiDAR & 0.495 & 0.164 & \textbf{0.402}$^{\dagger}$ & 0.433 \\
\bottomrule
\end{tabular}
}}
\vspace{-3mm}
\end{table}

\subsubsection{Real-World Fine-Tuning Results}
Table~\ref{tab:bfpp_real_world_main} reports the
held-out real-world comparison. LiDAR modestly improves overall and work-zone
F1, camera-only leads on lanes, and both tie on trajectories. Hence LiDAR data helps more in real-world than in simulation. (Supplementary
Sec.~S4).

\begin{table}[!t]
\centering
\caption{BF++ real-world fine-tuning with
one scenario held out for testing and the other three scenarios used
for fine-tuning.}
\vspace{-2mm}
\label{tab:bfpp_real_world_main}
{
\scriptsize
\setlength{\tabcolsep}{2.2pt}
\begin{tabular}{@{}lrrrr@{}}
\toprule
\textbf{Model} & \textbf{\makecell{Overall\\F1@0.5}} &
\textbf{\makecell{Lane\\F1@0.5}} &
\textbf{\makecell{Work-zone\\F1@0.5}} &
\textbf{\makecell{Trajectory\\F1@0.5}} \\
\midrule
BF++-Camera & 0.840 & \textbf{0.895} & 0.509 & \textbf{0.990} \\
BF++-Camera+LiDAR & \textbf{0.844} & 0.882 & \textbf{0.549} & \textbf{0.990} \\
\bottomrule
\end{tabular}
}
\vspace{-3mm}
\end{table}

\subsubsection{Closed-Loop and Efficiency Results}
Table~\ref{tab:bfpp_closed_loop_main} compares BF++
with SimLingo~\cite{Renz_2025_CVPR} and TransFuser++
(TF++)~\cite{Jaeger_2023_ICCV}. The benchmark now contains 76 scenarios and
228 weather-route evaluations. For a matched comparison, Table~\ref{tab:bfpp_closed_loop_main}
retains the original 180-route campaign and the 31 Town10 routes common to all
four methods at the evaluation freeze, giving 211 routes. BF++ outperforms both
baselines at real-time rates. Full Town10 results and the complete 228-route
audit are reported in Supplementary Sec.~S5.2, Table~S3, and Table~S-A2.

\begin{table}[!t]
\centering
\caption{Closed-loop and efficiency results. C/F/S are clear, fog, and storm
Driving Scores over the 211-route common comparison. Efficiency is measured
end to end on one NVIDIA RTX 5070 Ti.}
\label{tab:bfpp_closed_loop_main}
\vspace{-2mm}
{
\textbf{(a) Closed-loop performance}\\[0mm]
\setlength{\tabcolsep}{1.8pt}
\resizebox{\columnwidth}{!}{%
\begin{tabular}{lrrrrrrr}
\toprule
\textbf{Method} & \textbf{C} & \textbf{F} & \textbf{S} & \textbf{DS$\uparrow$} &
\textbf{RC$\uparrow$} & \textbf{SR$\uparrow$} & \textbf{Coll.$\downarrow$} \\
\midrule
BF++-Camera+LiDAR & \textbf{68.0} & 63.7 & \textbf{61.6} & \textbf{64.4} & 87.0 & 34.3 & 47 \\
BF++-Camera & 65.8 & \textbf{64.1} & 59.0 & 63.0 & \textbf{87.3} & 31.3 & \textbf{45} \\
SimLingo & 60.1 & 59.8 & 57.9 & 59.3 & 74.0 & \textbf{37.4} & 71 \\
TransFuser++ & 25.2 & 26.5 & 26.4 & 26.1 & 50.2 & 1.5 & ${\sim}244$ \\
\bottomrule
\end{tabular}
}

\vspace{1mm}
\textbf{(b) Efficiency}\\[0mm]
\setlength{\tabcolsep}{3.5pt}
\resizebox{\columnwidth}{!}{%
\begin{tabular}{lrrr}
\toprule
\textbf{Model} & \textbf{Parameters} & \textbf{Latency} & \textbf{FPS} \\
\midrule
BF++-Camera & 33.5M & 9.6 ms & \textbf{104.5} \\
BF++-Camera+LiDAR & 34.77M & 11.3 ms & 88.2 \\
SimLingo & 1,364.8M & 727 ms & 1.4 \\
TransFuser++ (3-model ens.) & 362M & 50.9 ms & 19.6 \\
\bottomrule
\end{tabular}
}
}
\vspace{-7mm}
\end{table}

\section{Conclusion}
\label{sec:conclusion}
\vspace{-2mm}
This work introduced \textbf{\textit{WorkZonePlan}}, a multimodal benchmark comprising 149,478 synthetic and 5,178 real-world samples with 3D lane, work-zone-boundary, and trajectory annotations, together with 76 CARLA work-zone scenarios and 228 weather-route evaluations for closed-loop testing. The accompanying \textbf{\textit{WAVE}} pipeline supports scalable generation and annotation of these data. We also proposed a novel network called \textbf{\textit{BoundaryFormer}} (\textbf{\textit{BF}}). The \textbf{BF} ablation shows that long-horizon trajectory reasoning benefits from a dedicated decoder rather than treating trajectories as interchangeable scene slots. Building on this result, \textbf{BF++} incorporates metric ground-plane representations, typed boundary and trajectory queries, long-range anchors, curve refinement, and optional gated LiDAR fusion. Across synthetic in- and out-of-distribution tests, held-out real-world evaluation, and closed-loop driving, BF++ provides structured perception and planning at 88.2--104.5 FPS with only 33.5--34.77M parameters. On the 211 routes common to all methods at the evaluation freeze, its camera and camera+LiDAR variants achieve closed-loop Driving Scores of 63.0 and 64.4, respectively, exceeding SimLingo and TransFuser++; the complete 228-route results in Supplementary Sec.~S5.2 preserve this ranking. Although broader out-of-distribution generalization remains an important direction for further study, these results demonstrate that jointly learning work-zone geometry and route-compatible trajectory options is a practical and promising research direction for end-to-end planning in work zones.

\vspace{-3mm}






    \small
    \bibliographystyle{ieeenat_fullname}
    \bibliography{root}

\clearpage
\setcounter{section}{0}
\setcounter{subsection}{0}
\setcounter{subsubsection}{0}
\setcounter{table}{0}
\setcounter{figure}{0}
\setcounter{equation}{0}
\renewcommand{\thesection}{S\arabic{section}}
\renewcommand{\thesubsection}{\thesection.\arabic{subsection}}
\renewcommand{\thesubsubsection}{\thesubsection.\arabic{subsubsection}}
\renewcommand{\thetable}{S\arabic{table}}
\renewcommand{\thefigure}{S\arabic{figure}}
\renewcommand{\theequation}{S\arabic{equation}}
\renewcommand{\theHsection}{S\arabic{section}}
\renewcommand{\theHsubsection}{S\arabic{section}.\arabic{subsection}}
\renewcommand{\theHsubsubsection}{S\arabic{section}.\arabic{subsection}.\arabic{subsubsection}}
\renewcommand{\theHtable}{S\arabic{table}}
\renewcommand{\theHfigure}{S\arabic{figure}}
\renewcommand{\theHequation}{S\arabic{equation}}

\begin{center}
{\LARGE\bfseries Supplementary Material}
\end{center}
\vspace{2mm}

\section{Original BF Objective and Implementation}
\label{supp:bf}

This supplement preserves the complete specification and results of the
original \textbf{BoundaryFormer (BF)} while separating them from its successor,
\textbf{BoundaryFormer++ (BF++)}. BF has two variants. In every section and
figure, \textbf{BF-IntegratedPlanner} denotes trajectory prediction from the
shared slot-attention heads, and \textbf{BF-DistinctPlanner} denotes the
dedicated trajectory decoder, with shared slots reserved for boundaries.

\subsection{Original BF architecture}
The original BF processes one RGB image with a pretrained Swin Transformer and
projects its spatial tokens into a shared latent space. Learnable slots attend
to those tokens and produce cubic coefficients, valid longitudinal range,
semantic type, and objectness for lane and work-zone boundaries. In
BF-IntegratedPlanner, those same slot heads also predict trajectory options. In
BF-DistinctPlanner, a separate global decoder fuses mean-pooled visual context
with objectness-weighted boundary-slot context and emits $K$ trajectory
hypotheses and mode probabilities. The latter design preserves object-level
boundary representations while giving long-horizon trajectory reasoning a
dedicated pathway.

\subsection{Multimodal trajectory options in BF}
The trajectory output is a candidate set, not a single curve
duplicated across slots. The annotation schema retains every locally valid
trajectory option, including both alternatives when a scene admits two. Let
$\widehat{\mathcal{T}}=\{(\widehat T_k,\pi_k)\}_{k=1}^{K}$ denote the predicted
curves and normalized mode probabilities. BF-IntegratedPlanner forms this set
from slots classified as trajectories; BF-DistinctPlanner produces it directly
with the dedicated decoder. In both variants, all options are generated from
perception before any route-level command is applied. A route planner may then
use command compatibility and $\pi_k$ to select one option, but it neither
changes the predicted geometry nor reveals the ground-truth best mode.

\subsection{Complete BF matching objective}
All BF outputs are jointly optimized with Hungarian assignment. Matching uses
semantic compatibility and geometric similarity between cubic coefficients and
valid spatial ranges. For each matched prediction--ground-truth pair, weighted
Smooth-$L_1$ losses supervise coefficients and range, cross-entropy supervises
type, and sampled points on the curves provide an alignment loss. Unmatched
predictions and ground-truth instances incur false-positive and false-negative
penalties. Objectness uses binary cross-entropy with logits. The full objective
is
\begin{equation}
\begin{aligned}
\mathcal{L}_{\mathrm{BF}}={}&\mathcal{L}_{\mathrm{coeff}}
+\mathcal{L}_{\mathrm{range}}+\mathcal{L}_{\mathrm{type}}
+\mathcal{L}_{\mathrm{curve}}\\
&+\mathcal{L}_{\mathrm{obj}}+\mathcal{L}_{\mathrm{FP}}
+\mathcal{L}_{\mathrm{FN}}+\lambda_{\mathrm{traj}}
\mathcal{L}_{\mathrm{traj}}.
\end{aligned}
\label{eq:supp_bf_loss}
\end{equation}
Here, $\mathcal{L}_{\mathrm{traj}}$ is active for BF-DistinctPlanner and
supervises the dedicated multimodal decoder. Trajectory
assignment is restricted to predicted and annotated trajectory sets, so a
trajectory option never competes with a lane or work-zone boundary in the
matching. Matched options supervise curve geometry and their mode scores;
unsupported options are penalized through confidence/objectness supervision.
This is the original BF objective; it is not reused to rename or retroactively
redefine BF++.

\section{Complete Original BF Results}
\label{supp:bf_results}

Table~\ref{tab:supp_bf_results} contains the threshold-wise BF-DistinctPlanner
results. Experiment~2
uses the original 137K Town01/Town10 training subset with the paper's 90/10 split;
Experiment~3 fine-tunes on all 5K real-world samples with the same 90/10 split.

\subsection{Open-loop evaluation metrics}
\label{supp:open_loop_metrics}
We apply standard detection statistics to distance-thresholded,
instance-level matching for lane boundaries, work-zone boundaries, and the BF
trajectory-option set. Each
predicted and annotated cubic curve $x=ay^3+by^2+cy+d$ is sampled at 50 points
over its overlapping longitudinal range. The mean 3D Euclidean point distance
is the pairwise cost. Evaluation greedily forms one-to-one matches in ascending
cost order at $\tau\in\{0.1,0.3,0.5\}$ m. A matched pair is a true positive
(TP), an unmatched prediction is a false positive (FP), and an unmatched
annotation is a false negative (FN). We compute
\begin{equation}
P=\frac{\mathrm{TP}}{\mathrm{TP}+\mathrm{FP}},\quad
R=\frac{\mathrm{TP}}{\mathrm{TP}+\mathrm{FN}},\quad
F1=\frac{2PR}{P+R},
\end{equation}
with F1@$0.5$ m primary. Mean absolute error (MAE) and root mean squared error
(RMSE) summarize the matched 3D curve distances at $0.5$ m, while the per-type
confusion matrix measures lane/work-zone/trajectory identity errors. For a
trajectory row, $P$ is option precision and $R$ is option coverage; the
multimodal and mode-selection metrics are defined in Sec.~S3.6. BF++ uses 38
metric anchors rather than 50 polynomial samples. {Its validation
summary also reports the total objective from
Eq.~\ref{eq:supp_bfpp_loss}, its objectness and point-regression terms, and
route-selected ADE over the 38 anchors. The corresponding partition-specific
results are reported in Tables~\ref{tab:bfpp_open_loop_loss_main}--\ref{tab:bfpp_real_world_main}.}

\begin{table*}[t]
\centering
\caption{Detection and localization performance of BF-DistinctPlanner across
the original experiments. Precision (P), Recall (R), and F1 are percentages at
$\tau\in\{0.1,0.3,0.5\}$ m. MAE and RMSE are in meters and use matched pairs at
$\tau=0.5$ m.}
\label{tab:supp_bf_results}
\small
\setlength{\tabcolsep}{3.5pt}
\begin{tabular}{@{}c@{\hspace{4pt}}l|ccc|ccc|ccc|cc@{}}
\toprule
\textbf{Experiment} & \textbf{Boundary Type} &
\multicolumn{3}{c|}{$\boldsymbol{\tau=0.1}$ m} &
\multicolumn{3}{c|}{$\boldsymbol{\tau=0.3}$ m} &
\multicolumn{3}{c|}{$\boldsymbol{\tau=0.5}$ m} &
\multicolumn{2}{c@{}}{\textbf{Loc. Error @ 0.5 m}} \\
& & P & R & F1 & P & R & F1 & P & R & F1 & MAE & RMSE \\
\midrule
\multirow{4}{*}{\textbf{Exp. 2}}
& Lane Boundary & 9.8 & 10.6 & 10.2 & 65.1 & 70.1 & 67.5 & 82.2 & 88.6 & 85.3 & 0.2258 & 0.2486 \\
& Work Zone Boundary & 9.9 & 8.9 & 9.4 & 41.6 & 37.5 & 39.5 & 58.8 & 53.1 & 55.8 & 0.2290 & 0.2612 \\
& Driving Trajectory & 47.5 & 46.6 & 47.0 & 68.3 & 66.9 & 67.6 & 76.7 & 75.2 & 75.9 & 0.1291 & 0.1711 \\
& \textbf{Overall} & 16.1 & 16.5 & 16.3 & 61.5 & 63.0 & 62.2 & 77.2 & 79.1 & 78.2 & 0.2095 & 0.2388 \\
\midrule
\multirow{4}{*}{\textbf{Exp. 3}}
& Lane Boundary & 26.5 & 28.4 & 27.4 & 76.0 & 81.4 & 78.6 & 89.8 & 96.2 & 92.9 & 0.1763 & 0.2091 \\
& Work Zone Boundary & 4.6 & 12.6 & 6.7 & 22.2 & 60.9 & 32.5 & 29.8 & 81.9 & 43.7 & 0.2167 & 0.2467 \\
& Driving Trajectory & 59.0 & 86.0 & 70.0 & 61.8 & 90.1 & 73.4 & 62.1 & 90.5 & 73.7 & 0.0419 & 0.0565 \\
& \textbf{Overall} & 24.4 & 37.0 & 29.4 & 52.0 & 78.8 & 62.6 & 60.7 & 91.9 & 73.1 & 0.1483 & 0.1864 \\
\bottomrule
\end{tabular}
\end{table*}

\begin{figure}[t]
\centering
\includegraphics[width=0.98\linewidth]{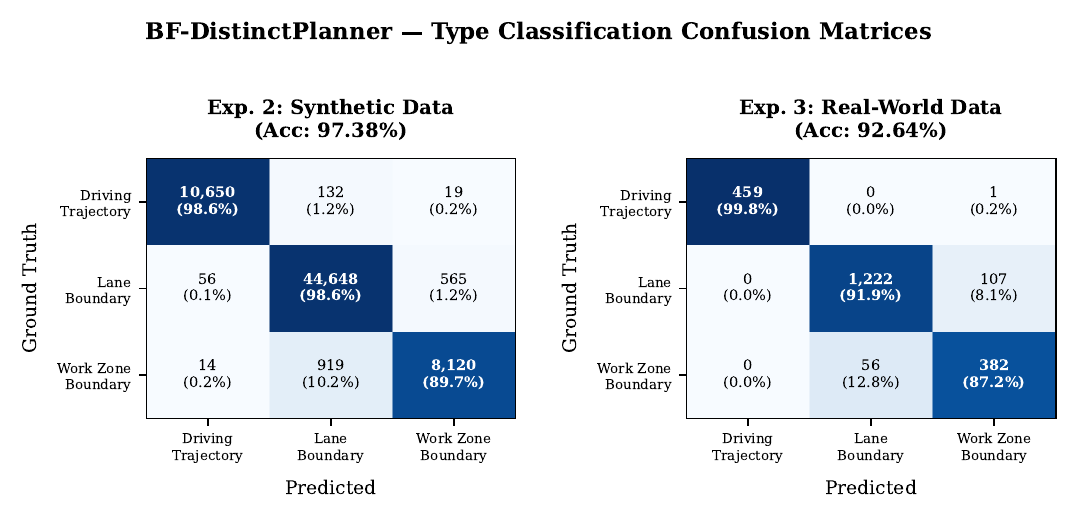}
\caption{Type-classification confusion matrices for BF-DistinctPlanner in
Experiments~2 and 3 at the 0.5 m distance threshold.}
\label{fig:supp_bf_confusion}
\end{figure}

\section{BF++ Architecture and Training}
\label{supp:bfpp}

\subsection{Design lineage}
BF++ follows the conclusion of the original BF ablation: trajectories require
an explicit global role rather than an interchangeable shared slot. Table
\ref{tab:bf_to_bfpp} summarizes the resulting redesign. The contribution is a
work-zone-specific set of geometric and topological inductive biases, not the
individual use of an existing backbone, attention layer, or matching routine.

\begin{table}[t]
\centering
\caption{From original BF to BF++.}
\label{tab:bf_to_bfpp}
\scriptsize
\setlength{\tabcolsep}{3.2pt}
\begin{tabularx}{\columnwidth}{@{}lXX@{}}
\toprule
\textbf{Property} & \textbf{BF} & \textbf{BF++} \\
\midrule
Coordinates & image-token latent space & calibrated metric ground plane \\
Entity roles & shared set slots & typed lane/WZ/trajectory queries \\
Geometry & cubic coefficients/range & 38 forward anchors, visibility/range \\
Trajectory & distinct global decoder & explicit typed query, two options \\
Evidence & one-pass slot decoding & two-stage curve reprojection/refinement \\
Modalities & camera & camera or gated camera+LiDAR \\
\bottomrule
\end{tabularx}
\end{table}

The physical representation follows the structure of a reconfigured road.
Typed lane and work-zone queries preserve the two sides of the temporary
drivable corridor; visibility and valid range allow curves to start and stop;
metric anchors preserve forward continuity as corridors shift, split, or merge;
and the typed trajectory output remains globally distinct from boundary
instances. These constraints prevent permutation-based matching from exchanging
physical roles that are not interchangeable in a work zone.

\subsection{BF++-Camera}
RGB is resized from $1920{\times}1080$ to $768{\times}432$ and normalized. An
ImageNet-pretrained ConvNeXt-Tiny produces stride-4/8/16/32 features. Stride-16
and stride-32 maps are projected by $1{\times}1$ convolution and GroupNorm to
$d=256$, then summed at $48{\times}27$. Camera intrinsics map every token to a
flat-ground forward/lateral coordinate used as metric positional encoding.
The CARLA camera has a $90^\circ$ field of view and is mounted at
$(1.6,0,1.7)$ m with $-15^\circ$ pitch.

A three-layer Transformer decoder ($d=256$, dropout 0.1) uses 17 typed queries.
{These comprise one trajectory query at zero lateral offset, four
anchored lane queries at $-6,-2,+2,+6$ m with a 2.5 m residual scale, six free
lane queries at $-9,-4.5,0,+4.5,+9,0$ m with a 9 m residual scale, and six
work-zone queries at $\pm1,\pm3,\pm5$ m. The repeated zero-offset free-lane
query is intentional and provides a second wide-residual center slot. The
dedicated trajectory query simultaneously emits two route-selectable
centerline options.}
Each option has independently regressed geometry and a confidence score; the
two curves share global query context but neither is generated conditionally
from the other. Each curve uses 38 forward anchors spanning 1.5--100 m. Heads
predict lateral offsets, per-anchor visibility, objectness, and a valid forward
range. Two refinement stages sample
stride-4/8/16 image evidence along the projected curve at heights
$\{0,0.5,1.0\}$ m and regress lateral and visibility corrections.

\subsection{BF++-Camera+LiDAR}
The perception LiDAR has 64 channels, vertical field of view
$+10^\circ/-30^\circ$, 100 m range, approximately 16K points/frame at 20 Hz,
and is co-located with the camera. A synchronized sweep is projected through
fixed LiDAR-to-camera extrinsics and z-buffered at $768{\times}432$ into three
channels: depth $\operatorname{clip}(d,0,60)/60$, height
$\operatorname{clip}((h+1)/4,0,1)$, and a hit mask. Four stride-2
$3{\times}3$ convolution--GroupNorm--GELU blocks of width
48/96/192/256 and a $3{\times}3$ head produce a $48{\times}27$ embedding.

Fusion is deliberately conservative:
\begin{equation}
\mathbf{f}\leftarrow\mathbf{f}+\mathbf{g}\odot E_{L}(\mathbf{r}),
\qquad \mathbf{g}\in\mathbb{R}^{256},\quad\mathbf{g}_0=\mathbf{0}.
\label{eq:supp_fusion}
\end{equation}
The zero gate makes the initial fused model identical to its camera parent, and
the learned correction is admitted channel by channel. Decoder, heads, and
refiner are otherwise unchanged. BF++-Camera has 33.5M parameters;
BF++-Camera+LiDAR has 34.77M, of which 1.24M belong to the LiDAR branch.

\subsection{Loss and optimization}
Ground-truth polynomials are sampled at the 38 metric anchors. Predictions are
matched independently within each entity type using masked mean point distance,
so lane, work-zone, and trajectory identities cannot exchange through matching.
\begin{samepage}
\par\noindent The BF++ objective is
{
\begin{equation}
\mathcal{L}_{\mathrm{BF++}}=5\mathcal{L}_{\mathrm{point}}
+\mathcal{L}_{\mathrm{vis}}+2\mathcal{L}_{\mathrm{obj}}
+0.5\mathcal{L}_{\mathrm{range}}+\mathcal{L}_{\mathrm{ref}},
\label{eq:supp_bfpp_loss}
\end{equation}
}
\end{samepage}
{
where $\mathcal{L}_{\mathrm{ref}}$ supervises the two-stage curve corrections.
The point term receives a further $2.5{\times}$ weight on work-zone curves.
The ``Point'' and ``Obj.'' entries in Table~\ref{tab:bfpp_open_loop_loss_main} are already
weighted contributions, rather than raw losses. Together with the visibility,
range, and refinement contributions, they sum to the reported total. For
example, the OOD camera row gives
$12.532+3.125+0.423+0.009+0.355=16.443$. The applied weights are 5.0 for point
regression, with the additional work-zone factor above, 2.0 for objectness,
1.0 for visibility, and 0.5 for range.
}
The fused
model initializes all camera weights from BF++-Camera. For epochs 0--2, the
camera trunk is frozen while the LiDAR branch and gate train at
$2\times10^{-4}$. For epochs 3--11, decoder, heads, and refiner are unfrozen at
one-quarter of that learning rate while the backbone remains frozen. We use
AdamW, weight decay 0.05, batch size 12, gradient clipping at 1.0, no geometric
augmentation, and Gaussian image noise $\sigma=0.02$. The fixed training and
validation partitions are summarized in Sec.~S6.

\subsection{Common route command, controller, and speed governor}
\label{supp:controller}
{
The typed trajectory query produces two options. At evaluation, every model
receives the same benchmark expected-route command (left/right/straight). Both
curves are decoded before handoff. The command masks incompatible options, and
the controller selects the highest-confidence remaining option. The command is
therefore analogous to a destination-level instruction. It is not an encoder
input, does not expose local boundary geometry, and cannot select the
lowest-error option after ground truth or execution is observed.

The selected centerline is tracked by pure pursuit. For lateral aim-point
offset $y_{\mathrm{aim}}$, lookahead $\ell=6$\,m, and wheelbase $L=2.85$\,m,
the steering angle and normalized command are
\begin{equation}
\begin{aligned}
\delta&=\operatorname{atan2}\!\left(2Ly_{\mathrm{aim}},
\ell^2+y_{\mathrm{aim}}^2\right),\\
\widetilde u_s&=\operatorname{clip}\!\left(\frac{\delta}{70^\circ},-0.7,0.7\right).
\end{aligned}
\label{eq:supp_steering}
\end{equation}
The command is exponentially smoothed with coefficient 0.6. When the predicted
left and right lane boundaries form a geometrically consistent adjacent pair,
lane snapping places the reference centerline at their midpoint. Lane snapping
is enabled in the deployed configuration but is suppressed within 25\,m of a
detected work zone, where the predicted work-zone trajectory is retained.

Both BF++ variants use the same separate 32-channel utility LiDAR for
longitudinal control. This sensor is distinct from the 64-channel perception
LiDAR available only to BF++-Camera+LiDAR. In the utility-LiDAR frame, $x$ is
forward, $y$ is right, and $z$ is up. The base filter retains points satisfying
$-2.2<z<0.4$\,m and $2<x<35$\,m. The center corridor further requires
$|y|\leq1.4$\,m. If it contains at least four points, the obstacle distance
$d$ is the fifth percentile of their forward ranges $x$. This statistic is
less sensitive to isolated returns than the absolute minimum. A narrow
emergency corridor uses $|y|\leq0.9$\,m and $x<20$\,m. Its distance
$d_{\mathrm{nar}}$ is likewise the fifth percentile when at least three points
are present.

The stopping-distance governor sets
\begin{equation}
v_{\mathrm{lim}}=\sqrt{\max\!\left(0,2a(d-d_{\mathrm{stop}})\right)},
\qquad
v_{\mathrm{des}}=\min(v_{\mathrm{nom}},v_{\mathrm{lim}}),
\label{eq:supp_speed_governor}
\end{equation}
with $a=2.0$\,m/s$^2$, $d_{\mathrm{stop}}=3.0$\,m, and deployed nominal speed
$v_{\mathrm{nom}}=8.9$\,m/s. All reported runs set
\texttt{BF\_TARGET\_SPEED=8.9}; the 11.2\,m/s code fallback was not used. The
narrow-corridor condition $d_{\mathrm{nar}}<2.4$\,m overrides
Eq.~\ref{eq:supp_speed_governor} and sets $v_{\mathrm{des}}=0$. Time to
collision is logged for analysis but does not enter the active speed limiter.

Let $e=v_{\mathrm{des}}-v$ denote speed error. Throttle $u_t$ and brake $u_b$
are generated by the custom rule
\begin{equation}
\begin{aligned}
u_t&=\begin{cases}
\operatorname{clip}(0.55e,0,0.75), & e\geq0,\\
0, & e<0,
\end{cases}\\[1mm]
u_b&=\begin{cases}
1, & v_{\mathrm{des}}<0.15\ \land\ v>0.3,\\
\operatorname{clip}(-0.6e,0,1), & e<-0.4,\\
0, & \text{otherwise}.
\end{cases}
\end{aligned}
\label{eq:supp_actuation}
\end{equation}
Thus, negative errors within 0.4\,m/s use a coast dead band. This controller is
neither a PID nor CARLA's built-in vehicle controller. {
BF++-Camera describes the learned model's input modality; its complete evaluated
driving stack additionally uses the non-learned utility-LiDAR speed governor.
Holding that governor fixed makes the two BF++ variants directly comparable.
It does not, by itself, isolate the contribution of the learned architecture
relative to SimLingo or TransFuser++, whose native speed-control systems differ.}
}

\subsection{Trajectory-option metrics}
\label{supp:trajectory_metrics}
Let
$\mathcal{T}^{*}=\{T^{*}_{j}\}_{j=1}^{M}$ be the annotated valid options and
$\widehat{\mathcal{T}}=\{(\widehat T_k,\pi_k)\}_{k=1}^{K}$ the predictions.
For the common visible anchor set $\mathcal{A}$, their mean point distance is
\begin{equation}
d(\widehat T_k,T^{*}_{j})=
\frac{1}{|\mathcal{A}|}\sum_{a\in\mathcal{A}}
\|\widehat T_k(a)-T^{*}_{j}(a)\|_2.
\end{equation}
If $\mathcal{M}_{\tau}$ is the one-to-one trajectory matching at threshold
$\tau$, option coverage and option precision are
\begin{equation}
\mathrm{Cov}@\tau=\frac{|\mathcal{M}_{\tau}|}{M},\qquad
\mathrm{Prec}_{\mathrm{opt}}@\tau=\frac{|\mathcal{M}_{\tau}|}{K}.
\end{equation}
Accordingly, recall and precision in the trajectory rows of
Table~\ref{tab:supp_bf_results} are exactly option coverage and option
precision, respectively; matched MAE/RMSE measure the geometry of recovered
options. We also define the oracle hypothesis diagnostic
\begin{equation}
\mathrm{minADE}=\frac{1}{M}\sum_{j=1}^{M}
\min_{k}d(\widehat T_k,T^{*}_{j}).
\end{equation}
This metric measures whether the candidate set contains a good option, but it
does not measure whether the model or route planner selects it. With
$k_{\mathrm{top1}}=\arg\max_k\pi_k$ and
$k_{\mathrm{route}}=\arg\max_{k\in\mathcal{C}(c)}\pi_k$, where
$\mathcal{C}(c)$ is the subset compatible with route command $c$, top-1 ADE and
route-selected ADE compare those respective curves with the command-compatible
annotation. BF++ reports route-selected ADE and executes that same mode in
closed loop; it never uses $\mathrm{minADE}$ or an oracle best-of-two mode for
the deployment claim.

{We additionally evaluated the frozen trajectory head on the fixed
2,791-frame validation split. Among frames with valid trajectory diagnostics,
the 2,745 single-option cases give a top-1 ADE of 0.081 m (median 0.025 m) and
an oracle minADE of 0.075 m. For the 35 multi-option cases, top-1 ADE is
0.246 m, oracle minADE is 0.124 m, and both annotated options are recovered
within 0.5 m in 30/35 cases. Mode-count accuracy is 98.7\% for single-option
frames and 100\% for multi-option frames. These oracle values diagnose the
coverage of the predicted set; they are not used for route handoff or
closed-loop scoring.}

\subsection{In-distribution open-loop analysis}
\label{supp:bfpp_id_open_loop}
The two final BF++ variants use the same fixed
2,791-frame Town01/Town10 validation split. Tables~\ref{tab:bfpp_open_loop_loss_main}
and~\ref{tab:bfpp_open_loop_f1_main} provide
the split-by-sensor comparison. Route-ADE is the mean error of the
command-selected trajectory over 38 anchors and is not an oracle best-of-two
result. The fused model slightly reduces total and point losses, while both
variants obtain the same Route-ADE and nearly identical overall F1. These
results indicate that open-loop performance is saturated on familiar towns.

\subsection{Additional open-loop evaluation}
\label{supp:bfpp_ood_open_loop}
{The OOD evaluation separates missed detections from localization
error. Lane F1 decreases from 0.97 in distribution to 0.50 OOD, while OOD
precision remains 0.72 and recall decreases to 0.52. Among matched curves, mean
point error increases more moderately, from 0.08 m to 0.20 m. Work-zone
detection is also sensitive to the confidence threshold learned in simulation:
on Town02, lowering the threshold from 0.30 to 0.03 raises work-zone F1 from
0.08 to 0.36 without changing the predicted curves. Town02 remains the hardest
split because its complex turns and geometry. LiDAR is
approximately neutral in the aggregate OOD open-loop results. This pattern is
consistent with the closed-loop failure analysis, in which 77\% of collisions
occur in Towns02 and 05.}

\section{BF++ Real-World Fine-Tuning Details}
\label{supp:bfpp_real_world}
{
\subsection{Physical sensors and fused model input}
The real subset is drawn from WorkZone3D. Its physical rig uses a telephoto
forward camera and six 16-channel VLP-16 LiDAR units. The released KITTI-format
data fuse the six units into one ego-registered point cloud per frame. BF++
therefore consumes the merged cloud, rather than a single physical 64-channel
sensor. A representative release frame contains approximately 43K points,
spans $-6.1^\circ$ to $+18.0^\circ$ in elevation, and occupies approximately
206 distinct $0.1^\circ$ elevation bins. This density and angular support are
consistent with the six differently mounted VLP-16 sensors.

The native forward RGB stream is $1024\times768$ with a
$25.3^\circ\times17.3^\circ$ field of view. Depending on the capture sequence,
camera/LiDAR acquisition is 30/20\,Hz or 10/10\,Hz. Precise timestamps and the
6-DoF UTM vehicle state support camera--LiDAR synchronization and alignment.

We content-matched the experiment images to the release's
\texttt{image\_1} camera and selected its calibration by direct
LiDAR-to-image reprojection. At the $768\times432$ model input, the recovered
intrinsics are $f_x=1747.3$, $f_y=1318.1$, $c_x=387.1$, and $c_y=235.3$.
The extrinsics place the camera 2.04 m forward and 1.14 m above ground with
$-1.62^\circ$ pitch. Projection through the release calibration places 653
points in the image on the expected lane markings, work-zone objects, and
vehicles, compared with 4 and 9 in-image points for the two alternative
calibrations. The same calibration is used during fine-tuning and evaluation.

The merged point cloud is projected using \texttt{P2} and
\texttt{Tr\_velo\_to\_cam}, then rasterized at $768\times432$. Its three input
channels are depth normalized by 60 m, height above ground over $[-1,3]$ m,
and a z-buffered hit mask. These channel definitions are identical to those
used for the simulated perception LiDAR.

\subsection{Leave-one-scenario-out protocol}
The raster-paired subset contains 5,178 released frames from four independent
capture sequences, which we denote Scenarios 1--4. The result reported in
Table~\ref{tab:bfpp_real_world_main} holds out Scenario 2, a nighttime sequence containing 773 frames,
for testing and uses Scenarios 1, 3, and 4 for fine-tuning. We fine-tune the
complete network for eight epochs with the backbone learning rate set to 0.1
times the head learning rate, batch size 12, AdamW, and weight decay 0.05.

\subsection{Annotation consistency audit}
We performed a read-only audit on 152 frames sampled at uniform temporal
intervals, with 38 frames from each scenario. Released lane, work-zone, and
trajectory labels were projected onto the corresponding RGB image using the
verified calibration. Every visible labeled entity was classified as correct,
requiring correction when the apparent offset exceeded approximately 0.3 m,
or spurious; visible but unlabeled entities were counted as missing. The audit
uses the exact labels from the reported experiments. No labels were changed
and no model was retrained.

The audit was a single-rater, AI-assisted screening in which a vision-capable
model inspected every projected overlay. Its judgments concern apparent image
alignment and should not be interpreted as centimetre-level 3D measurements.
Geometric support is claimed only where LiDAR supplies an independent
reference, such as work-zone-object clusters and the reprojection check above.

\begin{table}[t]
\centering
\caption*{{\textbf{Table S-A1.} Read-only consistency audit of
the real-world annotations. Counts are entity-level; ``correct'' is the
fraction requiring no correction.}}
\scriptsize
\setlength{\tabcolsep}{2.0pt}
\resizebox{\columnwidth}{!}{%
\begin{tabular}{lrrrrr}
\toprule
\textbf{Class} & \textbf{Inspected} & \textbf{Missing} & \textbf{Spurious} &
\textbf{Correction} & \textbf{Correct} \\
\midrule
Lane boundary & $\sim$470 & 0 & 0 & 0 & 100\% \\
Work-zone boundary & $\sim$135 & 1 & 0 & 8 & 93.3\% \\
Trajectory & $\sim$154 & 0 & 0 & 2 & 98.7\% \\
\bottomrule
\end{tabular}}
\end{table}

Overall, 143/152 frames (94.1\%) contain no audit flag. The nine flagged frames
cluster in two cases: ambiguous taper geometry in Scenario 1, where a labeled
closure boundary crosses lane dashes that appear open in the image, and
far-range polynomial-tail curvature beyond 15 m in Scenario 4. Near-field
alignment within 15 m is consistent across all sampled frames. In the fold
where Scenario 4 is held out, its weaker performance appears primarily
photometric, owing to darkness and glare, rather than the result of systematic
annotation displacement. Two sampled frames contain valid dual-trajectory
forks, and both alternatives are labeled.
}

\section{Closed-Loop Evaluation Details}
\label{supp:closed_loop_details}
\subsection{Protocol and metric aggregation}
\label{supp:closed_loop_metrics}
Closed-loop evaluation follows the CARLA/Bench2Drive
protocol. The benchmark contains 60 work-zone scenarios across CARLA
Towns01--05 and 16 additional in-distribution scenarios in Town10. Replaying
all 76 scenarios under clear-day, foggy-dusk, and storm-night conditions gives
228 scheduled routes per method. For route $r$,
\begin{equation}
\mathrm{RC}_r=100\frac{d_r^{\mathrm{completed}}}{d_r^{\mathrm{route}}},
\qquad
\mathrm{DS}_r=\mathrm{RC}_r\mathrm{IP}_r,
\end{equation}
where $\mathrm{IP}_r\in[0,1]$ is the accumulated infraction penalty. Success
Rate is the percentage of routes satisfying the benchmark's full-route
success condition, and the collision count is accumulated over the evaluated
routes. We report weather-specific and pooled results together with paired
route-level Driving Score comparisons. The
route command selects a compatible trajectory before rollout, as detailed in
Sec.~S3.5; no metric selects an option after observing ground truth or
execution outcomes.

{An agent termination caused by blocking, route deviation, or a
collision is treated as a completed evaluation and contributes its penalized
Driving Score to every aggregate. A simulator or harness failure produces no
valid endpoint score and is rerun under the same configuration and seed. No
route is excluded because of driving performance. At the evaluation freeze,
211 routes had valid results for all
four methods. After that freeze, the remaining
simulator or harness failures were rerun. The completed audit therefore
contains one valid endpoint for each of the $4\times228=912$ scheduled
method--route records.}

{
\begin{table*}[t]
\centering
\caption*{{\textbf{Table S-A2.} Complete 228-route closed-loop
results. Table~\ref{tab:bfpp_closed_loop_main} reports the 211 routes
common to all four methods at the evaluation freeze, whereas this table uses
every route after completing the interrupted evaluations.
Success (``Perfect'') is defined as DS$\geq99.9$; SR reports the exact count
and percentage.}}
\small
\setlength{\tabcolsep}{4.4pt}
\begin{tabular}{lcccc}
\toprule
\textbf{Metric} & \textbf{BF++-Camera+LiDAR} &
\textbf{BF++-Camera} & \textbf{SimLingo} & \textbf{TransFuser++} \\
\midrule
Valid records & 228/228 & 228/228 & 228/228 & 228/228 \\
Clear DS & \textbf{68.7} & 66.3 & 60.6 & 25.4 \\
Fog DS & \textbf{64.3} & 63.8 & 60.6 & 26.3 \\
Storm DS & \textbf{60.6} & 58.4 & 60.3 & 26.1 \\
Overall DS & \textbf{64.5} & 62.8 & 60.5 & 26.0 \\
RC (\%) & 87.1 & \textbf{87.5} & 74.1 & 49.4 \\
SR & 76/228 (33.3\%) & 69/228 (30.3\%) &
\textbf{87/228 (38.2\%)} & 3/228 (1.3\%) \\
Collisions & \textbf{47} & \textbf{47} & 71 & 261 \\
\bottomrule
\end{tabular}
\end{table*}

\begin{table*}[!t]
\centering
\caption*{{\textbf{Table S-A3.} Closed-loop scenario-family
breakdown on the pre-rerun 208-route audit subset. Each entry is Driving
Score / successes / collision events.}}
\small
\setlength{\tabcolsep}{4.2pt}
\begin{tabular}{lcccc}
\toprule
\textbf{Family ($n$ routes/model)} & \textbf{BF++-Camera+LiDAR} &
\textbf{BF++-Camera} & \textbf{SimLingo} & \textbf{TransFuser++} \\
\midrule
Merge--detour--return (81) & \textbf{63.7}/32/29 & 63.3/31/31 &
50.3/22/47 & 28.1/0/104 \\
Lane-change/merge (85) & 59.2/17/8 & 58.1/15/4 & \textbf{60.5}/34/11 &
24.6/0/82 \\
Town01 straight open lane (15) & \textbf{85.4}/11/3 & 82.5/9/2 &
73.6/10/4 & 23.3/0/16 \\
Town01 merge-left-return (12) & 81.4/8/0 & 70.5/7/0 &
\textbf{92.5}/8/1 & 14.3/0/26 \\
Shift-and-return/detour-like (9) & 44.9/0/7 & \textbf{47.2}/0/7 &
42.9/0/7 & 15.2/0/12 \\
Straight within lane (6) & \textbf{85.0}/3/0 & \textbf{85.0}/3/0 &
83.1/4/1 & 63.0/3/3 \\
\bottomrule
\end{tabular}
\end{table*}

Table~\ref{tab:bfpp_closed_loop_main} pools the original 180-route campaign with the 31 Town10
routes common to all methods at the evaluation freeze. Table~S-A2 instead
includes all 48 completed Town10 routes, giving 228 records per method.
BF++-Camera+LiDAR has the highest Driving Score in both comparisons, while
SimLingo has the highest Success Rate.}

\subsection{Town10 in-distribution extension}
\label{supp:town10_closed_loop}
Town10 adds 16 work-zone scenarios and 48 scheduled weather routes per model.
The interrupted evaluations were rerun under the same harness, configuration,
and seeds, giving 48/48 valid records for every method.
Table~\ref{tab:supp_town10_closed_loop} reports all 16 routes in each weather.

\begin{table}[t]
\centering
\caption{Town10 closed-loop results on all 48 routes for each method.
C/F/S are clear, fog, and storm Driving Scores.}
\label{tab:supp_town10_closed_loop}
\scriptsize
\setlength{\tabcolsep}{2.2pt}
\resizebox{\columnwidth}{!}{%
\begin{tabular}{lrrrrr}
\toprule
\textbf{Method} & \textbf{DS} & \textbf{RC} & \textbf{Coll.} &
\textbf{Perfect} & \textbf{C/F/S} \\
\midrule
BF++-Camera+LiDAR & \textbf{68.8} & \textbf{88.7} & \textbf{0} & 15/48 &
73.2/73.7/59.4 \\
BF++-Camera & 65.2 & 88.1 & 2 & 13/48 &
70.1/69.2/56.4 \\
SimLingo & 63.4 & 82.4 & 1 & \textbf{20/48} & 54.4/61.2/74.7 \\
TransFuser++ & 23.1 & 48.0 & 49 & 0/48 & 22.2/23.9/23.2 \\
\bottomrule
\end{tabular}}
\end{table}

BF++-Camera+LiDAR leads BF++-Camera by 3.6 Driving Score and records no
collisions on Town10. SimLingo has the highest perfect-route count, but lower
Driving Score and route completion than the fused BF++ model.
Table~\ref{tab:bfpp_closed_loop_main} uses the 31-route Town10 intersection available at the evaluation
freeze; Table~S-A2 pools all 48 completed records with the original 180 routes.

\subsection{Runtime and paired-route analysis}
Both BF++ variants were trained on one NVIDIA RTX
5090 and evaluated under the same deployment configuration on one NVIDIA RTX
5070 Ti. Runtime includes preprocessing and curve decoding. The LiDAR raster
and encoder add 1.7\,ms while retaining faster-than-real-time operation.
Parameter counts are exact checkpoint sums. TransFuser++ uses a
three-checkpoint ensemble. SimLingo uses
camera/IMU/GNSS, whereas TransFuser++ uses camera/LiDAR/IMU/GNSS. The separate
utility LiDAR is the common stopping governor for both BF++ variants and is not
a learned BF++-Camera input.

In the original Towns01--05 campaign, BF++-Camera+LiDAR changes paired-route
Driving Score by $+1.09$ relative to BF++-Camera, but the difference is not
significant ($t=1.09$). On the completed Town10 set, the difference is
$+3.6$ DS. {Across all 228 routes, BF++-Camera+LiDAR obtains 64.5
DS and BF++-Camera obtains 62.8. SimLingo and TransFuser++ obtain 60.5 and
26.0, respectively. The BF++ variants therefore record higher Driving Scores
and fewer collisions than the external baselines. SimLingo, however, has the
highest Success Rate, with 87/228 successes compared with 76/228 for
BF++-Camera+LiDAR.}

{
\subsection{External-baseline configuration and scope}
The evaluated checkpoints, sensors, navigation adapters, controllers, and
inference settings are summarized here. SimLingo uses the
authors' public \texttt{epoch=013} checkpoint (SHA-256 prefix
\texttt{ec894372}) with its native sensors and controller and is not fine-tuned
on WorkZonePlan. TransFuser++ uses the authors' public all-towns
three-checkpoint ensemble (prefix \texttt{d6fbdc28}), with all members evaluated
at each step. The BF++-Camera+LiDAR checkpoint has prefix \texttt{b47fd028};
the BF++-Camera control export has prefix \texttt{130eb363}. All methods receive
the same benchmark route command and are evaluated on the same machine and
harness with Epic rendering quality, a 600 s timeout, and one attempt per
route. Because the external methods retain their native speed controllers, the
fixed BF++ utility-LiDAR governor controls the camera-versus-fusion comparison
but does not independently isolate the learned architecture from controller
differences across model families.

\subsection{Scenario-family analysis and paired uncertainty}
We retain the pre-rerun 208-route audit subset for the previously computed
scenario-family analysis and group it using generator metadata when
available, including \texttt{route\_type} in Towns03--05 and the Town01 layout
variants. For remaining routes, a disclosed geometric rule assigns
shift-and-return paths with lateral displacement above 2.5 m to the detour-like
family, paths with net lateral displacement above 2.5 m to lane-change/merge,
and the remainder to straight. Table~S-A3 reports
Driving Score, integer successes, and collision events for each family. This
diagnostic subset is not used for the 211-route comparison or the
complete 228-route aggregate in Table~S-A2.

The largest gain occurs in the merge--detour--return family, where
BF++-Camera+LiDAR exceeds SimLingo by 13.4 Driving Score and records 29 rather
than 47 collisions. SimLingo is stronger on the clean lane-change/merge and
Town01 merge-left-return families, which also explains its higher aggregate
Success Rate.

We also perform a layout-blocked bootstrap with 10,000 resamples over 64
layouts. All weather replays from a layout are resampled as one unit.
BF++-Camera+LiDAR minus SimLingo has mean $\Delta$DS $=+5.04$, with 95\%
confidence interval $[-2.86,+12.98]$ and two-sided $p\simeq0.20$.
BF++-Camera minus SimLingo has mean $\Delta$DS $=+3.72$, interval
$[-3.92,+11.86]$, and $p\simeq0.35$. The paired differences are positive but
are not statistically significant at the layout level with 64 effective
units.

A qualitative inspection of an evaluated storm-night Town03 detour success
found that BF++-Camera+LiDAR completed the route with 100.0 DS while SimLingo
obtained 29.0 DS. In representative Town02 and Town04 failures, the agents left
the route before cone contact, consistent with the route-level OOD failure
pattern discussed in Sec.~S3.8.

}

\section{Town-Wise Dataset and Closed-Loop Distribution}
\label{supp:town_distribution}
Table~\ref{tab:supp_town_distribution} separates
released frames from closed-loop routes. {Following the accounting
used throughout the paper, the release contains 137,749 Town01/Town10
frames and 11,729 Town02--05 frames, giving 149,478 synthetic and 5,178 real
samples (154,656 total).
Towns01--05 contribute 12 closed-loop scenarios each, while Town10 contributes
16 additional in-distribution scenarios. Three weather replays per scenario
produce 228 scheduled routes from 76 base layouts.}

\begin{figure}[t]
\centering
\includegraphics[width=0.75\linewidth]{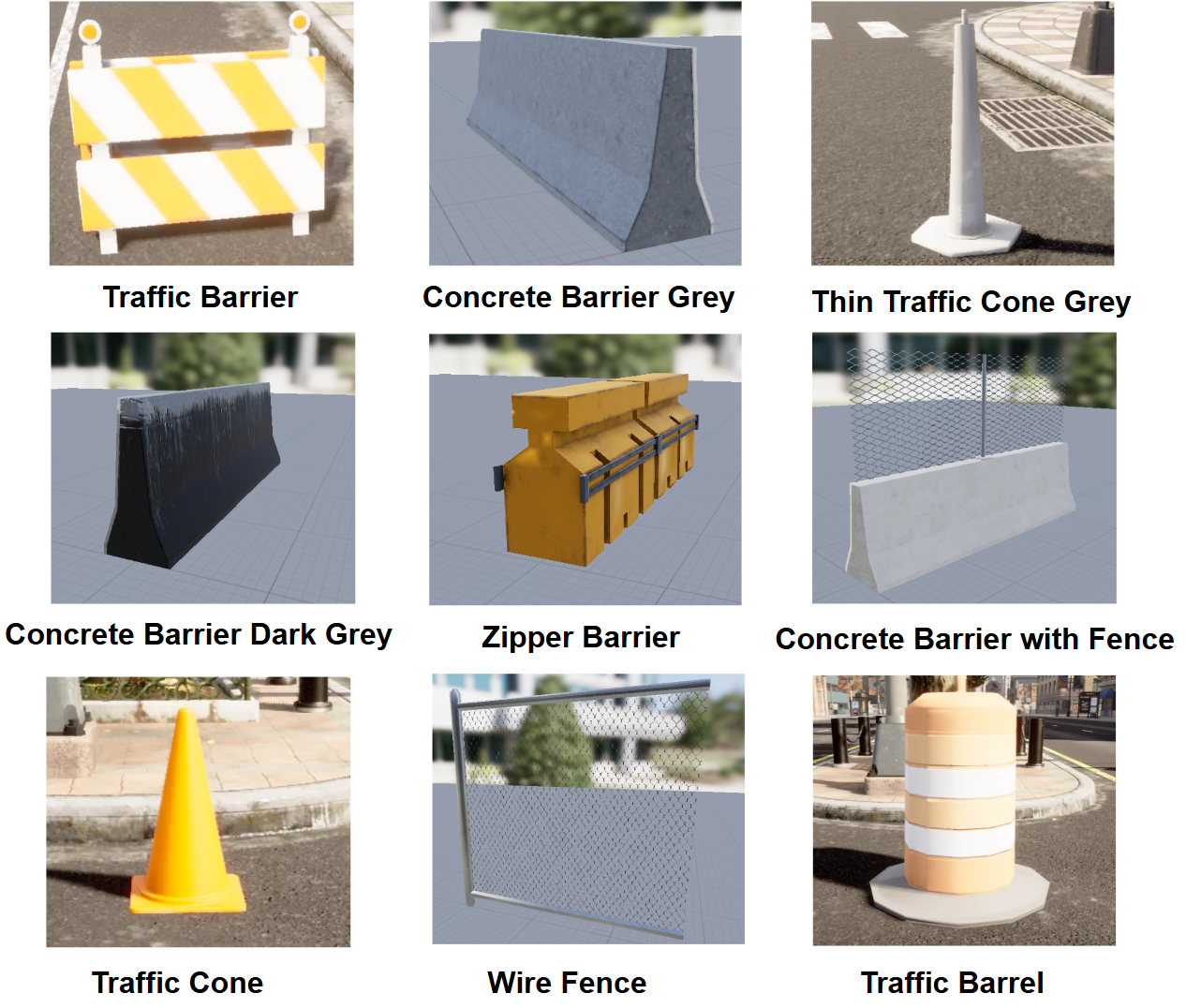}
\caption{The work-zone object classes in \textbf{\textit{WorkZonePlan}}.}
\label{fig:supp_syn_objects}
\end{figure}

The nine work-zone object classes used to construct the synthetic scenarios
are shown in Fig.~\ref{fig:supp_syn_objects}.

{
\begin{center}
\begin{minipage}{\columnwidth}
\captionof{table}{{Frame and closed-loop
distribution. Scen. counts base layouts; Routes counts their three weather
replays.}}
\label{tab:supp_town_distribution}
\centering
\scriptsize
\setlength{\tabcolsep}{2.4pt}
\renewcommand{\arraystretch}{1.04}
{
\begin{tabular}{lrlrr}
\toprule
\textbf{Town/source} & \textbf{Frames} & \textbf{Role} & \textbf{Scen.} &
\textbf{Routes} \\
\midrule
Town01 & 80K+ & ID corpus & 12 & 36 \\
Town10 & 50K+ & ID corpus & 16 & 48 \\
Reported ID subtotal & 137,749 & -- & 28 & 84 \\
Town02 & 2,797 & OOD holdout & 12 & 36 \\
Town03 & 4,945 & OOD holdout & 12 & 36 \\
Town04 & 1,792 & OOD holdout & 12 & 36 \\
Town05 & 2,195 & OOD holdout & 12 & 36 \\
\midrule
\textbf{Synthetic total} & \textbf{149,478} & -- & \textbf{76} &
\textbf{228} \\
WorkZone3D real & 5,178 & Real transfer & -- & -- \\
\textbf{Complete dataset} & \textbf{154,656} & -- & \textbf{76} &
\textbf{228} \\
\bottomrule
\end{tabular}}
\end{minipage}
\end{center}
}

{The BF++ experiments use 125,626 training frames and a fixed
2,791-frame validation subset drawn from the reported 137,749-frame
Town01/Town10 corpus. The remaining 9,332 ID frames are retained in the corpus
but are not used in the fixed BF++ train/validation comparison. The 11,729
Town02--05 frames are reserved for held-out evaluation and are added only when
forming the 149,478-frame synthetic release total. These counts define the
fixed training, validation, and held-out partitions used for the reported BF++
experiments.}

\end{document}